%% file: main.tex
\documentclass[letterpaper]{article} 
\usepackage[preprint]{aaai2027}
\usepackage[hyphens]{url}  
\usepackage{graphicx} 
\usepackage{natbib}  
\usepackage{caption} 
\input{meta/packages}
\input{meta/authors_arxiv}
\usepackage{algorithm}
\usepackage{algorithmic}
\usepackage{newfloat}
\usepackage{listings}
\DeclareCaptionStyle{ruled}{labelfont=normalfont,labelsep=colon,strut=off} 
\floatstyle{ruled}
\newfloat{listing}{tb}{lst}{}
\floatname{listing}{Listing}
\definecolor{paperlink}{HTML}{1F5F8B}
\usepackage[colorlinks=true,linkcolor=paperlink,citecolor=paperlink,urlcolor=paperlink]{hyperref}
\title{G-ReAct: Graph-Guided Deep Search via Structure-State Co-Evolution}

\begin{document}

\maketitle

\input{sections/0_Abs}


\input{sections/1_Intro}
\input{sections/2_Pre}
\input{sections/3_Method}
\input{sections/4_Exp}
\input{sections/5_Conclusion}

\setlength{\bibsep}{0.55em plus 0.1em}  
\bibliography{resources/references}
\input{sections/Appendix}


\end{document}

%% file: meta/packages.tex
\usepackage{amsmath,amsfonts,amssymb}
\usepackage{multirow}
\usepackage{booktabs}
\usepackage{colortbl}
\usepackage{fancyvrb}
\usepackage{tcolorbox}
\tcbuselibrary{listings,skins,breakable}
\usepackage{iftex}
\ifxetex
  \usepackage{fontspec}
  \IfFileExists{Inconsolatazi4-Regular.otf}{%
    \IfFileExists{Inconsolatazi4-Bold.otf}{%
      \setmonofont[
        Extension=.otf,
        UprightFont=Inconsolatazi4-Regular,
        BoldFont=Inconsolatazi4-Bold,
      ]{Inconsolatazi4}%
    }{}%
  }{}%
\fi

\renewcommand{\arraystretch}{1.12}                  

%% file: meta/authors_arxiv.tex
\newcommand{\authormark}[1]{%
\raisebox{0.8ex}{\scriptsize\ensuremath{\mathrm{#1}}}%
}

\author{
Shaoxiong Yang\authormark{1\mathord{*}\dagger}\quad
Mengyuan Zhang\authormark{1\dagger}\quad
Shaojun Lin\authormark{2\mathsection}\quad
Chao Li\authormark{1}\\
Wei Liu\authormark{1}\quad
Kun Shao\authormark{1\ddagger}\quad
Jian Luan\authormark{1\ddagger}
}

\affiliations{
\quad
\authormark{1}Xiaomi Inc.
\quad
\authormark{2}Huazhong University of Science and Technology\\
\texttt{yangshaoxiong007@126.com}\\
\texttt{\{zhangmengyuan7,lichao75,liuwei40,shaokun,luanjian\}@xiaomi.com}\\
\texttt{linshaojun@hust.edu.cn}
}

\makeatletter
\gdef\@thanks{%
\footnotetext[1]{~~Conducted during the author's employment at Xiaomi Inc.}%
\footnotetext[2]{~~Equal contribution.}%
\footnotetext[3]{~~Corresponding author.}%
\footnotetext[4]{~~Work done during internship.}%
}
\makeatother

%% file: sections/0_Abs.tex
\begin{abstract}
Deep search has become a fundamental capability of large language models (LLMs) for solving open-domain complex tasks. However, existing approaches typically rely on linear sequential reasoning for both trajectory generation and inference, making it difficult to consistently preserve intermediate states and constraints throughout long-horizon multi-hop search. Consequently, they often suffer from context forgetting, search drift, and inefficient exploration. To address these limitations, we propose \textbf{G-ReAct}, a reasoning framework for deep search that organizes reasoning as \textbf{state evolution over a fixed-topology query graph}. The evolving graph state explicitly tracks search progress and guides subsequent decisions, transforming exploratory search driven by textual history into graph-guided reasoning under explicit constraints. G-ReAct supports both training and inference: it generates high-quality deep-search trajectories for supervised fine-tuning and provides structured guidance for inference-time search without additional fine-tuning. Experiments demonstrate that with only 1.9K generated trajectories for fine-tuning, Qwen3-30B-A3B-Thinking-2507 achieves 52.6\% accuracy on BrowseComp-ZH and 79.0\% on XBench, outperforming comparable open-source methods trained on substantially larger datasets, including RL-enhanced methods. Furthermore, when applied at inference time, G-ReAct consistently improves the performance of existing strong LLMs on deep-search tasks. We will publicly release all code and model weights.
\end{abstract}

%% file: sections/1_Intro.tex
\section{Introduction}
LLMs have demonstrated remarkable capabilities in complex reasoning and knowledge-intensive tasks~\citep{wei2022cot}. However, their deployment in open-domain real-world scenarios remains limited by outdated parametric knowledge, factual hallucinations~\citep{huang2025hallucination}, and the difficulty of reliably leveraging external information~\citep{asai2024self}. To address these challenges, recent studies have developed \textbf{Deep Search Agents}~\citep{li2025search,jin2025searchr1,wu2025deepdive}, which augment LLMs with external tools such as search engines and web browsers to acquire up-to-date information through multi-step retrieval, verification, and reasoning. Compared with retrieval-augmented generation (RAG)~\citep{guu2020realm,lewis2020rag}, deep search requires models to continuously plan, explore, verify, and update reasoning states in dynamic open environments, making it a foundational capability for reliable open-domain agents.

\begin{figure}[!t]
\centering
\includegraphics[width=1\linewidth]{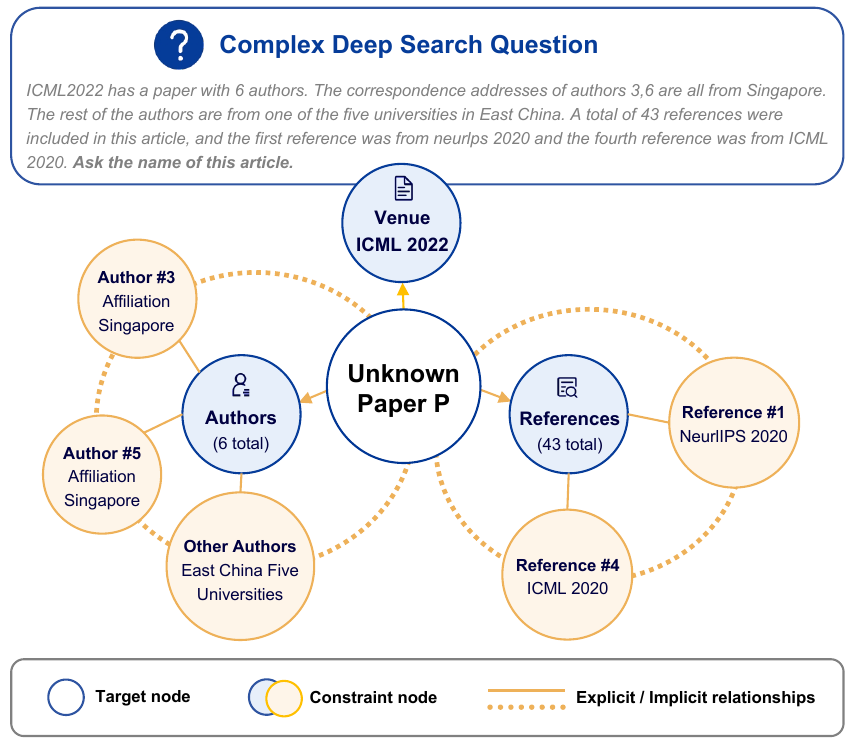}
\caption{An example of a deep-search problem requiring iterative search, evidence verification, and multi-step reasoning.}
\label{fig:example}
\end{figure}

To evaluate this capability, challenging and web-intensive benchmarks such as BrowseComp~\citep{wei2025browsecomp}, BrowseComp-ZH~\citep{zhou2025browsecomp}, GAIA~\citep{mialon2024gaia}, and XBench~\citep{xbench2025} have been proposed. As illustrated in Figure~\ref{fig:example}, these problems typically involve multi-hop dependencies, multiple unknown entities, implicit relations, and ambiguous constraints. Solving them requires models to progressively identify candidate entities, verify constraints, and complete the reasoning chain through multiple rounds of interaction. Therefore, deep search not only tests whether a model can effectively invoke external tools, but also depends critically on its ability to decompose problems, plan searches, integrate information, manage candidates, and maintain constraint consistency over long reasoning horizons.


Existing deep search agents are typically trained using challenging question-answer pairs together with their reasoning trajectories~\citep{li2025websailorv2,zheng2026redsearcher}. Knowledge graphs are widely used to construct difficult questions with multi-hop dependencies, ambiguous constraints, and verifiable answers, ensuring that each question is associated with a clear latent reasoning path~\citep{edge2024graphrag,guo2025falsecotqa}. However, compared with question generation, the trajectory construction and problem solving remain far less structured. Most existing methods rely on strong reasoning models to generate ReAct-style interaction trajectories~\citep{yao2023react}, followed by answer verification, format filtering, and process-level validation~\citep{miromind2026mirothinker,li2025openseeker}. Although ReAct provides a general and flexible ``thought -- action -- observation'' paradigm for tool-augmented reasoning, it still represents reasoning states primarily as free-form textual interaction history. During long-horizon deep search, models must simultaneously track multiple unknown entities, candidate sets, cross-step constraints, verified facts, and unresolved subgoals. Relying solely on textual context can easily lead to state dilution and constraint forgetting, resulting in redundant retrieval, constraint loss, and reasoning drift. This reveals a fundamental asymmetry: while knowledge graphs provide a clear logical backbone during question generation, such structural information is often weakened or even discarded during trajectory construction and problem solving, forcing models to rediscover the latent reasoning structure through trial and error.

Based on this observation, we argue that improving deep-search capability should not rely solely on increasing question difficulty or scaling reasoning trajectories. More importantly, it requires structured organization of reasoning states and decisions throughout the solving process. To this end, we propose \textbf{G-ReAct} (\textbf{Graph-guided Reasoning and Acting}), a co-evolutionary graph-structured reasoning framework for deep search. G-ReAct explicitly represents problem solving as a task-specific reasoning graph with a dual-layer design of \textbf{\emph{structure invariance}} and \textbf{\emph{state evolution}}. The structural layer represents the stable logical backbone induced by problem constraints, where nodes denote target entities or key concepts and edges capture their dependency relations. The state layer is attached to nodes and edges to record dynamically evolving search information, including candidate entities, verified facts, and constraint satisfaction states. During reasoning, the model gathers observations through external tools and continuously incorporates reliable atomic facts into the graph state. The evolving graph state then guides subsequent subgoal selection and search directions, forming a closed loop of ``retrieval -- verification -- update -- guidance'' In this way, G-ReAct transforms unstructured search driven primarily by textual interaction history into graph-directed reasoning under explicit structural constraints, thereby improving the coherence, stability, and interpretability of long-horizon search.


Our main contributions are summarized as follows:
\begin{itemize}
    \item We propose \textbf{G-ReAct}, a graph-structured reasoning framework for deep search that formulates deep search as state evolution under explicit structural constraints, supporting both trajectory construction during training and search guidance during inference.

    \item We introduce a \textbf{graph-reasoning co-evolution mechanism} that maintains candidate entities, verified facts, and constraint satisfaction states through a closed loop of ``retrieval -- verification -- graph update -- decision guidance'', thereby alleviating redundant retrieval, constraint loss, and reasoning drift in long-horizon search.

    \item Experiments show that \textbf{G-ReAct improves both search effectiveness and search efficiency}. Using only 1.9K generated trajectories for supervised fine-tuning, G-ReAct boosts Qwen3-30B-A3B-Thinking-2507 to 52.6\% on BrowseComp-ZH and 79.0\% on XBench-DS, surpassing comparable open-source methods trained on far larger datasets. As an inference-time framework, it further improves the performance of existing strong LLMs while requiring fewer search steps.
\end{itemize}

%% file: sections/2_Pre.tex
\section{Related Work}

\subsection{Deep Search Agents}
Deep search requires models to iteratively invoke external tools (e.g., Google Search) to acquire information, verify evidence, and refine search strategies over multiple interaction rounds. While closed-source systems such as OpenAI Deep Research~\citep{openai2025deepresearch}, Gemini Deep Research~\citep{google2025gemini}, and Claude Research~\citep{claude4} have demonstrated impressive capabilities, their training procedures remain undisclosed, making it difficult to reproduce their performance. Recent open-source research has broadly explored deep search agents from three complementary perspectives: training data construction, training paradigms, and inference frameworks, with representative efforts including MiroThinker~\citep{lu2025mirothinker} and OpenSeeker~\citep{li2025openseeker}.
Among these directions, training data construction has received the most systematic investigation. For QA construction, WebSailor~\citep{li2025websailor} pioneered the integration of knowledge graphs into deep search by generating multi-hop questions via random walks over Wikidata~\citep{vrandecic2014wikidata}. DeepDive~\citep{wu2025deepdive}, WebLeaper~\citep{tao2025webleaper}, and OpenSeeker further refined this pipeline through improved entity sampling, path expansion, and difficulty calibration. For reasoning trajectory construction, existing work primarily improves supervision quality via stronger teacher models, trajectory filtering~\citep{tao2025webleaper}, process verification~\citep{lu2025mirothinker}, and context compression, yet leaves the underlying representation of reasoning states largely unchanged. Different from these efforts, our work focuses on the organization of reasoning states, using an explicit graph structure to represent and maintain intermediate reasoning states for subsequent search decisions.

\subsection{Knowledge Graph-Guided Structured Reasoning}
Knowledge graphs organize structured knowledge as entities and relations, providing explicit relational constraints and structural priors for complex reasoning. Existing work mainly falls into two categories. One line exploits graph structures to support retrieval augmentation~\citep{guu2020realm} and reasoning organization during problem solving. Representative methods such as GraphRAG~\citep{edge2024graphrag} and Hyper-RAG~\citep{he2025hyperrag} model entity associations with graphs or hypergraphs to improve information organization, retrieval, and reasoning. Another line leverages knowledge graphs to construct challenging multi-hop question-answering datasets. For example, DeepDive~\citep{wu2025deepdive} and WebDancer~\citep{li2025webdancer} generate reasoning paths through controlled random walks, while FalseCoTQA~\citep{guo2025falsecotqa} builds adversarial multi-hop benchmarks to evaluate reasoning robustness.

Existing methods effectively exploit graph structures for retrieval augmentation, reasoning organization, and data construction. However, they primarily use graphs as static knowledge resources or offline organizational tools, rather than for explicit state maintenance and decision support during problem solving. In contrast, our work focuses on the dynamic role of graph structures throughout deep search, using them to represent intermediate reasoning states and guide subsequent search decisions.

%% file: sections/3_Method.tex
\section{Methodology}
\label{sec:method}

\begin{figure*}[t]
    \centering
        \includegraphics[width=1\textwidth]{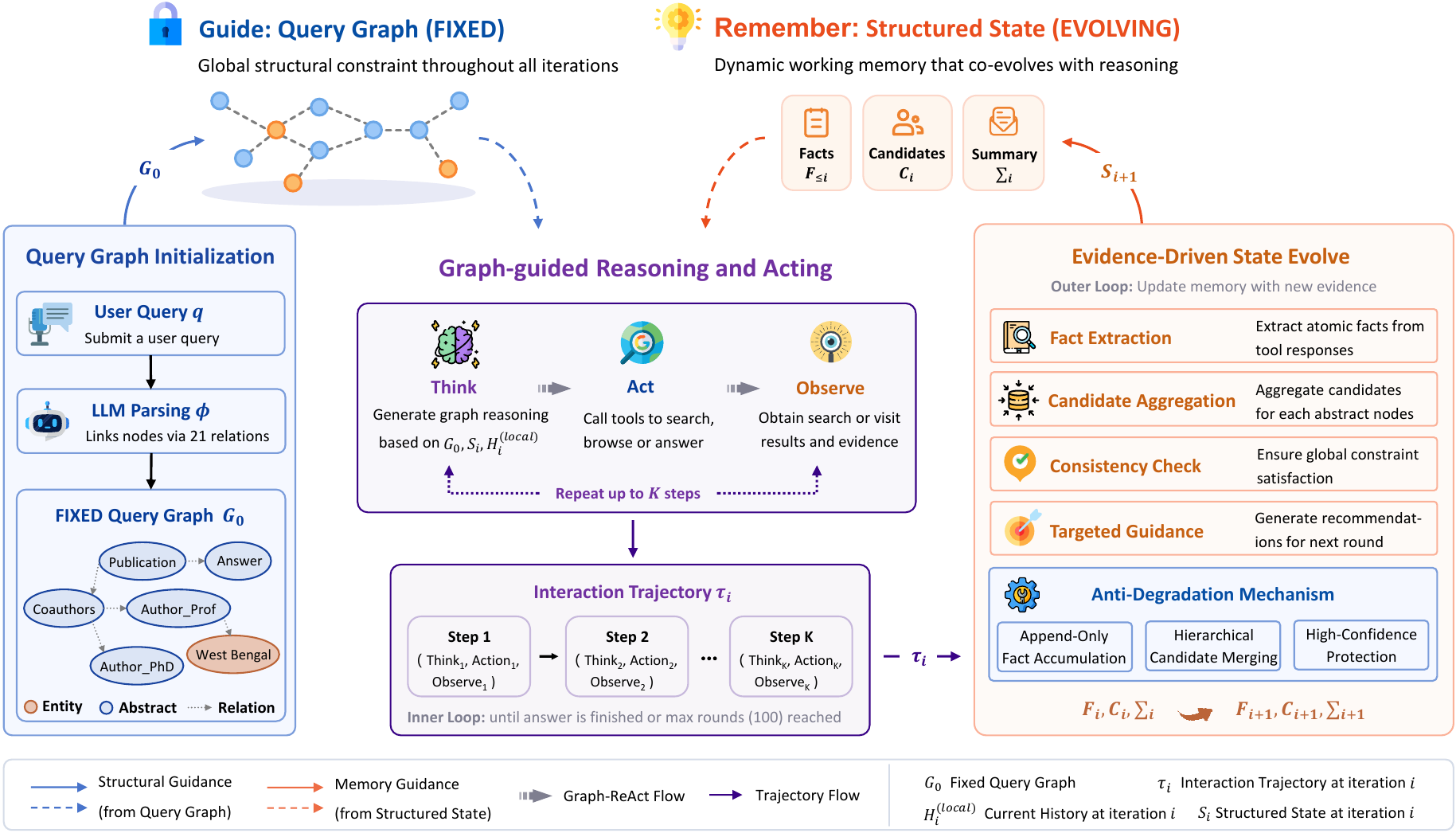}
    \caption{Overview of the G-ReAct framework, where a fixed query graph and an evolving structured state co-evolve to guide multi-round reasoning. Given a question, the initializer $\phi$ constructs a fixed-topology $G_0$ as the global constraint scaffold; within each round, the agent performs ``Think -- Act -- Observe'' exploration under their joint guidance; upon termination, an evidence-driven update extracts verified atomic facts, refreshes candidate sets and the consistency state, and feeds $S_{i+1}$ into the next round, realizing the ``structure-invariant, state-evolving'' design.}
    \label{fig:framework}
\end{figure*}

As illustrated in Figure~\ref{fig:framework}, \textbf{G-ReAct} structures long-horizon deep search over a fixed-topology query graph that stays structurally invariant while its attached reasoning state evolves across rounds, realizing a ``structure-invariant, state-evolving'' design. The framework proceeds in four stages: (1) \textbf{Query Graph Initialization}; (2) \textbf{Graph-guided Reasoning and Acting}, tool-augmented exploration over the graph; (3) \textbf{Evidence-Driven State Evolve}; and (4) \textbf{Iterative Refinement and Termination}.

\subsection{Query Graph Initialization}
\label{sec:query_graph_init}

We first parse the input question~$q$ into a query graph~$G_0$ that defines the constraint state space for subsequent reasoning,
\begin{equation}
\label{eq:graph_init}
G_0 = \phi(q) = (V_0,\, E_0),
\end{equation}
where $V_0=\{v_i\}_{i=1}^{|V|}$ and $E_0=\{e_j\}_{j=1}^{|E|}$ are the node and edge sets. The initializer~$\phi$ is a single, tool-free LLM call mapping~$q$ to a JSON graph over the schema below.

\textbf{Node Semantics.} Each node $v_i$ has type \texttt{abstract} or \texttt{entity}: \emph{abstract} nodes are reasoning slots resolved through multi-hop reasoning (e.g., \texttt{Answer}, \texttt{Author\_1}), while \emph{entity} nodes are named entities in~$q$ whose identifiers must match the original text verbatim. 


\begin{sloppypar}
\textbf{Relation Schema.} The relation set $\mathcal{R}=\{r_1,\ldots,r_{21}\}$ spans four families: compositional, attributional, usage, and constraint, with \texttt{associated\_with} as a fallback when no precise semantic match applies (full definitions in Appendix~F). Each edge $e_j=(v_s,v_t,r,\text{val},\text{anchor})$ has $r\in\mathcal{R}$, a value descriptor $\text{val}$, and $\text{anchor}\subseteq \mathrm{span}(q)$.
\end{sloppypar}

Beyond the schema, a strict \emph{grounding} principle requires every structural element of $G_0$ to be traceable to a span in~$q$; external information, synonym rewriting, and normalization are prohibited:
\begin{equation}
\label{eq:faithfulness}
\forall x\in V_0\cup E_0,\ \mathrm{anchor}(x)\subseteq\mathrm{span}(q).
\end{equation}
Here $\mathrm{span}(q)$ denotes the contiguous substrings of~$q$, and $\mathrm{anchor}(x)$ the verbatim span to which element~$x$ is grounded. Adhering to this principle ensures $G_0$ faithfully encodes the question's constraint structure as a grounded, verifiable graph rather than injecting external knowledge. The complete schema and relation ontology are detailed in Appendix~F.

\subsection{Graph-guided Reasoning and Acting}
\label{sec:g_react}

After obtaining $G_0$, the agent enters an exploration loop, as illustrated in the middle of Figure~\ref{fig:framework}: guided jointly by the fixed graph and the evolving state, it iteratively invokes external tools to collect evidence and validate graph constraints, producing a trajectory $\tau_i$ that drives the state evolve (\S\ref{sec:state_update}).

\subsubsection{Limitations of Sequential Reasoning Paradigms.}
\label{sec:sequential_limitations}

The standard ReAct method~\citep{yao2023react} conditions each decision on the flat interaction history $H_t = \langle q, (r_s, a_s, o_s)_{s=1}^{t} \rangle$, where $r_s, a_s, o_s$ are the reasoning trace, tool action, and observation at step $s$. As the sole state representation, this flat history leaves the structural dependencies and constraint propagation of multi-hop reasoning unmodeled. As the horizon grows, three systematic failures emerge: (i) \emph{redundant retrieval}: previously answered sub-questions are re-searched due to lack of progress tracking; (ii) \emph{constraint loss}: multi-hop constraints dissipate in long contexts~\citep{retrac2025}, so satisfying one constraint inadvertently violates another; and (iii) \emph{reasoning drift}: noise in the flat history progressively drives the agent off its original search objective~\citep{liu2024lost}.

\subsubsection{Graph-Constrained State-Space Decisions.}
\label{sec:graph_constrained_decisions}

To address these failures, G-ReAct reformulates reasoning as a \emph{graph-constrained state-space decision process} in which a fixed query graph $G_0$ and an evolving structured state $S_i$ jointly guide reasoning, restoring the progress tracking, constraint propagation, and objective anchoring that the flat history loses.

\textbf{Structured Reasoning State.} The structured state at iteration $i$ is
\begin{equation}
\label{eq:state}
S_i = (F_{\leq i},\; \mathcal{C}_i,\; \Sigma_i),
\end{equation}
with components
\begin{itemize}
    \item $F_{\leq i} = \bigcup_{s=1}^{i} F_s$: the cumulative set of verified facts;
    \item $\mathcal{C}_i$: the node candidate domain, containing candidate entity lists with confidence ratings for each abstract node;
    \item $\Sigma_i$: the global consistency state, recording reasoning progress, constraint satisfaction, and contradiction detection.
\end{itemize}
The initial state is $S_0 = (\emptyset, \emptyset, \emptyset)$.

\textbf{Graph-Conditioned Policy.} Unlike standard ReAct, which conditions on the entire flat history $H_t$, G-ReAct conditions each decision on the fixed graph $G_0$, the structured state $S_{i-1}$, and only the local interaction history of the current round:
\begin{equation}
\label{eq:policy}
a_t \sim \pi_{\text{G-ReAct}}\!\left(\,\cdot \mid G_0,\, S_{i-1},\, H_t^{(\mathrm{local})}\right),
\end{equation}
where $H_t^{(\mathrm{local})}$ denotes the tool interactions within the current exploration round, discarded at round end. Cross-round coherence is thus carried by $S_{i-1}$ rather than a growing textual context, avoiding the noise accumulation and history dilution of flat-history policies.

\textbf{Adaptive Search Strategy.} Beyond the state $S_{i-1}$, the agent decides \emph{where} to search next based on the confidence $\gamma$ of the current best candidate set in $\mathcal{C}_i$, adopting a three-tier strategy: \textbf{high} (verify remaining constraints to confirm the leading candidate), \textbf{medium} (limited verification, then broaden exploration on failure or contradiction), and \textbf{low} (discard candidates and explore graph-guided new directions), adaptively allocating exploration effort by evidence strength.

At the start of each round, $S_{i-1}$ is serialized into a structured text prompt that re-injects the question $q$, the fixed graph $G_0$, and the evolving state components (progress summary from $\Sigma_{i-1}$, top-$k$ candidates per node from $\mathcal{C}_{i-1}$ by confidence, and verified facts $F_{\leq i-1}$ annotated to prevent redundant re-searching), guiding the agent under both structural and state constraints. The rendering template is given in Appendix~G, and the tool specifications in Appendix~D.

\subsection{Evidence-Driven State Evolve}
\label{sec:state_update}

When a round terminates without producing a final answer, G-ReAct enters the state update phase, formalized as a transition function $\Psi$:
\begin{equation}
\label{eq:transition}
S_{i+1} = \Psi(S_i, \tau_i) = (F_{i+1},\; \mathcal{C}_{i+1},\; \Sigma_{i+1}),
\end{equation}
where $\tau_i = \{(r_{i,s}, a_{i,s}, o_{i,s})\}_{s=1}^{T_i}$ is the complete interaction trajectory of round $i$. $G_0$ remains fixed; $\Psi$ is a single,  tool-free LLM call that executes four structured analysis steps, each updating one component of $S_{i+1}$:

\textbf{Step 1: Atomic Fact Extraction.} We extract atomic-level facts $\{f_k\}$ from the tool responses, where each fact must satisfy \emph{atomicity} (a single proposition), \emph{traceability} (linked to a specific tool response), and \emph{purity} (no causal inference or evaluative judgment).

\textbf{Step 2: Candidate Entity Aggregation.} Next, we map the extracted facts $\{f_k\}$ to the corresponding nodes in $G_0$, aggregating new candidate entities for each abstract node $v$:
\begin{equation}
\label{eq:candidate}
\mathcal{C}_{i}^{(\mathrm{new})}(v) = \{(c_m,\, \mathrm{sup}_m,\, \mathrm{vio}_m,\, \gamma_m)\},
\end{equation}
where $c_m$ is a candidate entity grounded in the tool responses, $\mathrm{sup}_m$ and $\mathrm{vio}_m$ are the supporting and violating constraint sets (corresponding to satisfied/unsatisfied edges in $G_0$), and $\gamma_m \in \{\text{high}, \text{medium}, \text{low}\}$ is a confidence rating strictly determined by the proportion of satisfied constraints. $\mathcal{C}_{i+1}$ is then obtained by merging $\mathcal{C}_i^{(\mathrm{new})}$ with $\mathcal{C}_i$ under the anti-degeneration rules below.

\textbf{Step 3: Global Consistency Check.} Given $G_0$, $F_{\leq i}$, and $\mathcal{C}_i$, we check whether a candidate assignment satisfies all graph constraints, producing four signals in $\Sigma_{i+1}$: $\Sigma_{i+1}^{\mathrm{miss}}$ (uncovered constraints), $\Sigma_{i+1}^{\mathrm{best}}$ (the best assignment so far), $\Sigma_{i+1}^{\mathrm{cands}}$ (cross-node candidate combinations), and $\Sigma_{i+1}^{\mathrm{sat}}$ (a Boolean flag indicating full satisfaction). When $\Sigma_{i+1}^{\mathrm{sat}}=\text{true}$, the agent emits the answer; otherwise, $\Sigma_{i+1}^{\mathrm{miss}}$ guides the next search.

\textbf{Step 4: Targeted Exploration Guidance.} Based on the consistency check, we generate targeted recommendations for the next exploration round: which constraints remain unverified, which candidates require disambiguation, and which reasoning directions should be prioritized or abandoned.

Since $S_{i+1}$ aggregates across rounds, later rounds may lose earlier evidence. We address this with three anti-degeneration mechanisms that guarantee monotonic state progress.

\subsubsection{Anti-Degeneration Mechanisms and Monotonic State Progress.}
\label{sec:anti_degeneration}

We enforce three mechanisms:
\begin{enumerate}
\renewcommand{\labelenumi}{(\roman{enumi})}
\item \textbf{Append-only fact accumulation:} verified facts are never removed,
\begin{equation}
\label{eq:append_only}
\begin{split}
F_{i+1} = F_i \cup \big\{f \in F_i^{(\mathrm{new})} \,\big|\, \mathrm{fact}(f) \\
\notin \{\,\mathrm{fact}(g) \mid g \in F_i\,\}\big\}.
\end{split}
\end{equation}
\item \textbf{Hierarchical candidate merging:} recurrent candidates take the maximal confidence with unioned evidence,
\begin{equation}
\label{eq:merge}
\begin{split}
\gamma_m^{(i+1)} &= \max\!\big(\gamma_m^{(i)},\, \gamma_m^{(\mathrm{new})}\big), \\
\mathrm{sup}_m^{(i+1)} &= \mathrm{sup}_m^{(i)} \cup \mathrm{sup}_m^{(\mathrm{new})}.
\end{split}
\end{equation}
\item \textbf{High-confidence candidate protection:} any high-confidence candidate missing in the new round is re-injected,
\begin{equation}
\label{eq:protect}
\begin{split}
\forall v,c:&\ \gamma_c^{(i)} = \mathrm{high}\ \wedge\ c\notin\mathcal{C}_{i+1}(v) \\
&\Rightarrow\ \mathcal{C}_{i+1}(v)\leftarrow\mathcal{C}_{i+1}(v)\cup\{c\}.
\end{split}
\end{equation}
\end{enumerate}

\paragraph{Proposition 1 (Monotonic State Progress).}
\textit{Under the grounding constraint of $G_0$ (Eq.~\ref{eq:faithfulness}), the G-ReAct reasoning state is monotonically non-decreasing: for any $i < j$, (a) $F_i \subseteq F_j$, and (b) $\gamma_c^{(i)} \leq \gamma_c^{(j)}$ for all candidates $c$ present in both rounds. Moreover, (c) any high-confidence candidate appearing in round $i$ is preserved in all subsequent rounds. Since verifiable facts for a given question are finite and $\gamma_c$ takes values in a bounded set, the state converges to a fixed point.} (Proof in Appendix~C.)

This guarantee bounds the state evolution but does not ensure $G_0$ itself is correct, which depends on the initializer $\phi$; the grounding principle (Eq.~\ref{eq:faithfulness}) mitigates this risk by requiring every structural element to be traceable to the question text.

Sequential reasoning paradigms lack such a guarantee: with no append-only record of verified facts or protected candidates, progress is not monotone and earlier findings can be overwritten as the flat history grows.

\subsection{Iterative Refinement and Termination}
\label{sec:iterative_refinement}

G-ReAct employs a nested iteration architecture that balances exploration thoroughness against computational cost:
\begin{enumerate}
\item \textbf{Outer Loop (Cross-Round Refinement).} G-ReAct runs at most $K$ refinement iterations. At each round, $S_{i+1}$ is serialized into a fresh prompt (Appendix~G) and fed to a new G-ReAct exploration, while the full interaction history of the previous round is discarded: only the structured state persists. This state-only propagation transmits reasoning progress without accumulating raw text, avoiding the context degradation of long-sequence histories~\citep{trivedi2023interleaving} while preserving all verified facts.
\item \textbf{Inner Loop (Single-Round Exploration).} Within each round, the agent takes bounded tool calls until one of four conditions triggers: \emph{success} (a final answer satisfying all $G_0$ constraints), \emph{context overflow} (prompt length nears the context window), \emph{timeout} (the time budget is exceeded), or \emph{budget exhaustion} (the call limit is reached). Parameter settings are in Appendix~D.
\item \textbf{Forced Answer Generation.} If all $K$ iterations complete without convergence, the agent emits a best-effort answer from $F_{\leq K}$ and $\mathcal{C}_K$, selecting the candidate assignment that maximizes constraint satisfaction over $G_0$. The full procedure is provided in Appendix~A.
\end{enumerate}


%% file: sections/4_Exp.tex
\section{Experiments}
\label{sec:experiments}

\subsection{Experimental Setup}
\label{sec:exp_setup}

\label{sec:main_results}
\begin{table*}[!t]
\centering
\setlength{\tabcolsep}{3.5pt}
\renewcommand{\arraystretch}{1.05}
\small
\begin{tabular}{@{}lcccccc@{}}
\toprule
Model & Samples & Recipe & BrowseComp & BrowseComp-ZH & XBench-DS & GAIA \\
\midrule

\rowcolor{gray!8}
\multicolumn{7}{@{}l}{\textit{Closed-Source Models}} \\

Claude-4.5-Sonnet        & --- & --- & 24.1 & 42.4 & --- & 66.0 \\
Claude-4-Opus            & --- & --- & 18.8 & 37.4 & --- & --- \\
OpenAI-o3                & --- & --- & 49.1 & 68.7$^{\ddagger}$ & --- & 70.5 \\
OpenAI Deep Research     & --- & --- & 51.5 & 42.9 & --- & 67.4 \\
GPT-5-Thinking-High      & --- & --- & 54.9$^{\ddagger}$ & 63.0 & --- & 76.7$^{\ddagger}$ \\

\midrule

\rowcolor{gray!8}
\multicolumn{7}{@{}l}{\textit{Open-Source Large Models (230B--671B)}} \\

Kimi-K2-Instruct-1T      & --- & --- & 14.1 & 28.8 & --- & 57.7 \\
DeepSeek-V3.1-671B       & --- & --- & 30.0 & 49.2 & 71.2 & 63.1 \\
DeepSeek-V3.2-671B       & --- & --- & 51.4 & 65.0 & --- & 63.5 \\
GLM-4.6-357B             & --- & --- & 45.1 & 49.5 & --- & 71.9 \\
GLM-4.7-357B             & --- & --- & 52.0 & 66.6 & --- & --- \\
MiniMax-M2-230B          & --- & --- & 44.0 & 48.5 & --- & 75.7 \\

\midrule

\rowcolor{gray!8}
\multicolumn{7}{@{}l}{\textit{Open-Source Deep-Search Models ($\sim$30B)}} \\

WebDancer-32B            & ---   & SFT + RL      & 3.8  & 18.0 & 39.0 & 51.5 \\
MiroThinker-32B          & 147k  & SFT + RL      & 13.0 & 17.0 & ---  & 64.1 \\
WebSailor-32B            & ---   & SFT + RL      & 10.5 & 25.5 & 53.3 & 53.2 \\
WebSailor-V2-30B         & ---   & SFT + RL      & 35.3 & 44.1 & 73.7 & \textbf{74.1} \\
WebLeaper-30B            & 15k   & SFT         & 27.7 & ---  & 66.0 & 67.0 \\
Tongyi DeepResearch      & ---   & CPT + SFT + RL  & \textbf{43.4} & 46.7 & 75.0 & 70.9 \\
DeepDive-32B             & 4.1k  & SFT + RL      & 15.3 & 29.7 & 51.8 & --- \\

DeepDive-30B-A3B         & 0.6k    & SFT         & 12.7 & 21.1 & 44.0 & --- \\
\rowcolor{blue!5}
\textbf{G-ReAct-DeepDive (Ours)}
                          & 0.6k   & SFT         & 28.5 & 45.3 & 70.0 & --- \\

OpenSeeker-v1-30B-A3B    & 11.7k & SFT         & 29.5 & 48.4 & 74.0 & --- \\
\rowcolor{blue!5}
\textbf{G-ReAct-OpenSeeker-v1 (Ours)}
                          & 1.9k  & SFT         & 35.6 & \textbf{52.6} & \textbf{79.0}$^{\ddagger}$ & 64.2 \\

\bottomrule
\end{tabular}
\caption{
Main results on four deep-search benchmarks (pass@1, \%).
``Samples'' denotes the number of supervised training samples and ``Recipe'' the training recipe. Best results among $\sim$30B deep-search agents are shown in bold, and the best reported result on each benchmark is marked with $^{\ddagger}$.
``---'' indicates unreported results. The DeepDive/OpenSeeker-v1 comparisons are controlled experiments using the same underlying QA pools, isolating the effect of graph-structured trajectories.}
\label{tab:main}
\end{table*}

\paragraph{Benchmarks \& Metrics.}
We evaluate G-ReAct on four deep search benchmarks: \textbf{BrowseComp}~\citep{wei2025browsecomp}, \textbf{BrowseComp-ZH}~\citep{zhou2025browsecomp}, \textbf{GAIA}~\citep{mialon2024gaia}, and \textbf{XBench-DeepSearch} (XBench-DS)~\citep{xbench2025}. We use the complete test set of each benchmark, except for GAIA, where we adopt its 103-sample text-only subset~\citep{li2025search}. All results are reported as pass@1, judged by an LLM-as-a-judge (GPT-4o-mini)~\citep{zheng2023judging}, with inference at temperature 0.6 and top-$p$ 0.95. Detailed benchmark descriptions are provided in Appendix~E.

\paragraph{Baselines.}
We compare against three categories of models. \emph{Closed-source models}: Claude-4.5-Sonnet~\citep{anthropic2025sonnet45} and Claude-4-Opus~\citep{claude4}, OpenAI-o3~\citep{o3}, OpenAI Deep Research~\citep{openai2025deepresearch}, and GPT-5-Thinking-High~\citep{openai2025gpt5}. \emph{Open-source large models (230B--671B)}: Kimi-K2~\citep{kimi-k2}, DeepSeek-V3.1~\citep{deepseekv3.1} and DeepSeek-V3.2~\citep{deepseekv3.2}, GLM-4.6~\citep{glm2025glm46} and GLM-4.7~\citep{glm2025glm47}, and MiniMax-M2~\citep{minimax2025m2}. \emph{Open-source Deep-Search models} at the same $\sim$30B scale: WebDancer~\citep{li2025webdancer}, MiroThinker~\citep{lu2025mirothinker}, DeepDive~\citep{wu2025deepdive}, WebSailor~\citep{li2025websailor} and WebSailor-V2~\citep{li2025websailorv2}, WebLeaper~\citep{tao2025webleaper}, OpenSeeker-v1-SFT~\citep{li2025openseeker}, and Tongyi DeepResearch~\citep{li2025tongyi}; covering SFT, SFT+RL, and CPT+SFT+RL paradigms with data scales from 0.6k to 147k. Unless otherwise specified, baseline results are taken from official technical reports or public leaderboards.

\paragraph{Implementation Details.}
G-ReAct is built on Qwen3-30B-A3B-Thinking-2507~\citep{yang2025qwen3} and trained with SFT only, on 2$\times$8 A100 80GB GPUs with the Megatron framework. Training is conducted for 3 epochs with a batch size of 32, a learning rate of 1e-5, and a 128K-token context window. Full training configurations and the data-construction pipeline are provided in Appendix~D. To isolate the effect of graph-structured trajectories, we construct two controlled variants using the same underlying QA pools as their corresponding baselines, such that the primary difference lies in trajectory structure. Specifically, \textbf{G-ReAct-OpenSeeker-v1} is built from 1,898 QA pairs sampled from OpenSeeker-v1, and \textbf{G-ReAct-DeepDive} from 618 QA pairs sampled from DeepDive. The sampled QA pairs are then reprocessed with the G-ReAct pipeline to generate graph-structured trajectories.

\subsection{Main Results}
Table~\ref{tab:main} and Table~\ref{tab:inference} present the performance of G-ReAct in trajectory construction and inference guidance, respectively. We distill four key findings from these results.

\paragraph{(1) Performance: A 30B model establishes a new SOTA on XBench-DS with over 20$\times$ fewer parameters.} G-ReAct-OpenSeeker-v1 achieves \textbf{79.0} on XBench-DS, establishing a new state of the art among our compared baselines and leading DeepSeek-V3.1-671B by 7.8 points. On BrowseComp-ZH, it achieves \textbf{52.6}, surpassing GLM-4.6-357B (49.5). On BrowseComp, it reaches \textbf{35.6}, the best result among $\sim$30B open-source deep-search models. On GAIA, G-ReAct achieves 64.2, consistently outperforming WebDancer, WebSailor, and MiroThinker. Overall, a 30B model matches or even surpasses 600B+ open-source models and several frontier closed-source models across multiple benchmarks, indicating that parameter scale is not the sole determinant of deep-search performance; the organization of the reasoning process is equally important.

\paragraph{(2) Data Efficiency: Graph-structured trajectories enable highly efficient supervision.} In the same-source comparison, G-ReAct-DeepDive outperforms DeepDive-30B-A3B by \textbf{15.8}, \textbf{24.2}, and \textbf{26.0} points on BrowseComp, BrowseComp-ZH, and XBench-DS, respectively. Likewise, G-ReAct-OpenSeeker-v1 uses only \textbf{16\%} of the training trajectories (1.9K vs.\ 11.7K), yet still improves over OpenSeeker-v1-SFT by \textbf{6.1}, \textbf{4.2}, and \textbf{5.0} points on the same benchmarks. Beyond same-source comparisons, G-ReAct further surpasses WebSailor-V2 (two-stage pipeline), Tongyi DeepResearch (three-stage pipeline), and MiroThinker (147K trajectories, 77$\times$ more data). These results suggest that graph-structured trajectories carry substantially richer supervisory signals, reducing the reliance on large-scale training data and reinforcement learning.

\paragraph{(3) Inference-Time Guidance: Structured guidance improves both effectiveness and search efficiency.} Table~\ref{tab:inference} shows that, as a fine-tuning-free inference-time framework, G-ReAct consistently improves the performance of strong LLMs, boosting doubao-seed-2.0-pro by \textbf{6.57} points (64.71$\rightarrow$71.28), Claude-Sonnet-4.5 by \textbf{4.84} points (39.10$\rightarrow$43.94), and OpenAI-o3 by \textbf{2.54} points (54.33$\rightarrow$56.87). More importantly, these improvements are accompanied with \emph{fewer} tool calls—average reductions of 1.10, 1.41, and 1.49, respectively. The simultaneous increase in accuracy and decrease in search steps suggests that G-ReAct improves search efficiency rather than trading accuracy for more exploration. This is consistent with the design of G-ReAct, where the evolving graph state explicitly tracks search progress and guides subsequent decisions, suppressing redundant trial-and-error exploration and directing search toward unresolved graph states.

\paragraph{(4) Paradigm Implication: Structured reasoning alleviates the inherent limitations of linear ReAct at the representation level.} In deep search, redundant retrieval, constraint loss, and reasoning drift largely stem from the sequential trajectory representation adopted by ReAct, where evolving search states are implicitly accumulated in an ever-growing interaction history. Existing approaches mainly address these challenges by strengthening model capabilities through improved training and scaling. In contrast, G-ReAct explicitly maintains search progress and constraints through graph state evolution, reorganizing the reasoning process at the representation level. This also explains why G-ReAct achieves competitive performance without large-scale data or sophisticated training pipelines. We argue that for deep search, the design of reasoning substrates should be considered a fundamental dimension of progress, alongside improving model capabilities.

\begin{table}[t]
\centering
\setlength{\tabcolsep}{7pt}
\renewcommand{\arraystretch}{1.15}
\small
\begin{tabular}{lcc}
\toprule
Model & Pass@1$\uparrow$ & Avg.\ Tool Calls$\downarrow$ \\
\midrule
doubao-seed-2.0-pro & 64.71 & 19.72 \\
\rowcolor{blue!5}
\quad + G-ReAct & \textbf{71.28} & \textbf{18.62} \\

Claude-4.5-Sonnet & 39.10 & 24.32 \\
\rowcolor{blue!5}
\quad + G-ReAct & \textbf{43.94} & \textbf{22.91} \\

OpenAI-o3 & 54.33 & 18.71 \\
\rowcolor{blue!5}
\quad + G-ReAct & \textbf{56.87} & \textbf{17.22} \\
\bottomrule
\end{tabular}
\caption{Inference-time performance of G-ReAct on BrowseComp-ZH without fine-tuning. Lower Avg.\ Tool Calls indicate more efficient search.
}
\label{tab:inference}
\end{table}

\subsection{Ablation Studies}
\label{sec:ablation}

Table~\ref{tab:main} and Table~\ref{tab:inference} have demonstrated the generalizability of G-ReAct across diverse QA-pair distributions and strong backbone models. This section further investigates its designs from two perspectives: component contributions and design choices.

\paragraph{Component Contribution Analysis.}
Table~\ref{tab:ablation_component} evaluates the two core components in G-ReAct: \textbf{Query Graph} and \textbf{State Evolution}, through individual removal. Removing the query graph causes a larger drop on XBench than on BrowseComp-ZH, indicating its importance for complex information aggregation tasks. Removing state evolution leads to consistent degradation on both benchmarks, highlighting its role in maintaining intermediate states and reducing reasoning drift during long-horizon search. Both components provide complementary contributions to G-ReAct.

\paragraph{Design Choice Analysis.}
We further examine two key design choices: the state summarizer and the state-update granularity. Replacing the state summarizer with a stronger model does not improve performance on either benchmark, suggesting that the trained 30B base model has already internalized the summarization capability (Appendix~B). For state-update granularity, we compare four sliding-window sizes (32K / 64K / 96K / 128K), with 64K achieving the best performance across all metrics  (Table~\ref{tab:ablation_update}). This reflects a trade-off between reasoning continuity and timely state consolidation: updating too frequently (32K) fragments the reasoning chain, whereas consolidating too rarely (128K) lets the context degrade before the state is refreshed. Together, these analyses support the effectiveness of the current design choices in G-ReAct.

\begin{table}[t]
\centering
\setlength{\tabcolsep}{5pt}
\renewcommand{\arraystretch}{1.1}
\footnotesize
\begin{tabular}{@{}lcccc@{}}
\toprule
Variant &
\multicolumn{2}{c}{XBench} &
\multicolumn{2}{c}{BrowseComp-ZH} \\
\cmidrule(lr){2-3}\cmidrule(lr){4-5}
& Avg.\ p@1$\uparrow$ & p@3$\uparrow$
& Avg.\ p@1$\uparrow$ & p@3$\uparrow$ \\
\midrule
G-ReAct
& \textbf{75.00} & \textbf{90.00}
& \textbf{52.25} & \textbf{66.78} \\
\midrule
w/o Query Graph
& 67.00 & 80.00
& 48.44 & 64.01 \\
w/o State Evolution
& 69.67 & 84.00
& 46.88 & 61.25 \\
\bottomrule
\end{tabular}
\caption{
Component ablation on XBench and BrowseComp-ZH. Each ablated variant removes one component from G-ReAct. Best results in each column are shown in bold.
}
\label{tab:ablation_component}
\end{table}

\begin{table}[t]
\centering
\setlength{\tabcolsep}{4pt}
\renewcommand{\arraystretch}{1.1}
\footnotesize
\begin{tabular}{@{}lccc@{}}
\toprule
Window Size & Avg p@1$\uparrow$ & p@3$\uparrow$ & Ans.\ Rate$\uparrow$ \\
\midrule
32k   & 70.00 & 82.00 & 93.67 \\
\cellcolor{blue!5}64k   & \cellcolor{blue!5}\textbf{75.00} & \cellcolor{blue!5}\textbf{90.00} & \cellcolor{blue!5}\textbf{98.33} \\
96k   & 70.67 & 85.00 & 96.33 \\
128k  & 67.67 & 82.00 & 94.00 \\
\bottomrule
\end{tabular}
\caption{State-update granularity on XBench, all else identical. Best bolded.}
\label{tab:ablation_update}
\end{table}

%% file: sections/5_Conclusion.tex
\section{Conclusion}
\label{sec:conclusion}
We propose G-ReAct, a graph-structured reasoning framework for deep search that organizes reasoning as state evolution over a fixed-topology query graph. Through a dual-layer design of structure invariance and state evolution, G-ReAct supports the construction of high-quality structured reasoning trajectories during training and provides explicit structured guidance during inference, achieving consistent improvements across multiple deep-search benchmarks with markedly higher data efficiency. We hope this work provides a new perspective on structured reasoning for deep search and inspires future research on long-horizon reasoning more broadly.

%% file: sections/Appendix.tex
\appendix

\definecolor{appendixboxbg}{HTML}{F1F7F8}
\definecolor{appendixboxframe}{HTML}{7FA6B5}
\definecolor{appendixboxtitle}{HTML}{32677B}
\colorlet{appendixcodebg}{appendixboxbg}
\colorlet{appendixcodeframe}{appendixboxframe}

\lstdefinestyle{appendixcode}{
  backgroundcolor=\color{appendixcodebg},
  frame=none,
  framerule=0pt,
  framesep=0pt,
  rulecolor=\color{appendixcodeframe},
  xleftmargin=0pt,
  xrightmargin=0pt
}
\lstdefinestyle{appendixplaincode}{
  backgroundcolor=\color{appendixboxbg},
  frame=none,
  framerule=0pt,
  framesep=0pt,
  rulecolor=\color{appendixboxbg},
  xleftmargin=0pt,
  xrightmargin=0pt
}

\input{sections/AppendixA_Algorithm}      
\input{sections/AppendixB_Ablation}        
\input{sections/AppendixC_Proof}           
\input{sections/AppendixD_Implementation}  
\input{sections/AppendixE_Benchmarks}       
\input{sections/AppendixF_Schema}          
\input{sections/AppendixG_Prompts}         
\input{sections/AppendixH_CaseStudy}       

%% file: sections/AppendixA_Algorithm.tex
\section{G-ReAct Algorithm}
\label{app:algorithm}

Algorithm~\ref{alg:coe_graph} summarizes the full G-ReAct trajectory generation procedure, integrating query-graph initialization, graph-conditioned exploration, evidence-driven state update, and iterative refinement.

\begin{algorithm}[!h]
\caption{G-ReAct Trajectory Generation}
\label{alg:coe_graph}
\begin{algorithmic}[1]
\REQUIRE Question $q$; LLM $\mathcal{M}$; max rounds $K$; max calls $N_{\max}$ per round; context limit $L_{\max}$.
\STATE Initialize query graph $G_0 \leftarrow \phi(q)$ \COMMENT{graph initialization (Eq.~1, main text)}
\STATE Initialize state $S_0 \leftarrow (\emptyset, \emptyset, \emptyset)$
\FOR{$i = 1$ \TO $K$}
    \STATE Render structured prompt from $S_{i-1}$
    \STATE $H_i^{(\mathrm{local})} \leftarrow \emptyset$
    \FOR{$t = 1$ \TO $N_{\max}$}
        \STATE Sample $(r_t, a_t) \sim \pi_{\text{G-ReAct}}(\cdot \mid G_0, S_{i-1}, H_t^{(\mathrm{local})})$ \COMMENT{the policy (Eq.~4, main text)}
        \STATE $o_t \leftarrow \text{Env}(a_t)$; append $(r_t, a_t, o_t)$ to $H_t^{(\mathrm{local})}$
        \IF{final answer emitted and all $G_0$ constraints satisfied}
            \STATE \textbf{return} answer
        \ELSIF{context length $> L_{\max}$ \OR budget exhausted}
            \STATE \textbf{break} to state update
        \ENDIF
    \ENDFOR
    \STATE $S_i \leftarrow \Psi(S_{i-1}, \tau_i)$ \COMMENT{state transition (Eq.~5) and anti-degradation (Eqs.~7--9), main text}
    \IF{$\Sigma_i^{\mathrm{sat}}$}
        \STATE Emit final answer from $\Sigma_i.\texttt{best\_candidate\_set}$; \textbf{return}
    \ENDIF
\ENDFOR
\STATE Emit forced best-effort answer from $F_{\leq K}$, $\mathcal{C}_K$
\end{algorithmic}
\end{algorithm}

%% file: sections/AppendixB_Ablation.tex
\section{Ablation Studies}
\label{app:ablation}

This appendix details the state-summarizer ablation referenced in Section~4(Table\ref{tab:ablation_summarizer}).

\subsection{State-Summarizer Ablation}
\label{app:component_ablation}
By default, the evidence-driven state evolve $\Psi$ is performed by G-ReAct's fine-tuned model. We test whether a stronger, external model is better suited to this role by replacing it with \emph{doubao-seed} (configuration in Table~\ref{tab:ablation_summarizer}) while holding the rest of the pipeline fixed. As Table~\ref{tab:ablation_summarizer} shows, the swap leaves average p@1 essentially unchanged on XBench ($+0.33$) but reduces it by $3.29$ points on BrowseComp-ZH, and lowers p@3 on both benchmarks. This indicates that G-ReAct's training already internalizes the summarization capability: the fine-tuned model suffices, and a stronger generic model can harm performance through domain mismatch.

\begin{table}[!ht]
\centering
\caption{State-summarizer ablation. \emph{Own model} = G-ReAct's fine-tuned model used as the state summarizer; \emph{doubao-seed} = doubao-seed-1.8-251228 via API (thinking enabled, \texttt{reasoning\_effort = minimal}, 128k context, temperature 0.6, top-$p$ 0.95). Average p@1 over 3 runs; BC-ZH = BrowseComp-ZH. Best per column in \textbf{bold}.}
\label{tab:ablation_summarizer}
\setlength{\tabcolsep}{5pt}
\renewcommand{\arraystretch}{1.1}
\footnotesize
\begin{tabular}{@{}lcccc@{}}
\toprule
\multirow{2}{*}{Summarizer} & \multicolumn{2}{c}{XBench} & \multicolumn{2}{c}{BC-ZH} \\
\cmidrule(lr){2-3}\cmidrule(lr){4-5}
 & p@1$\uparrow$ & p@3$\uparrow$ & p@1$\uparrow$ & p@3$\uparrow$ \\
\midrule
\rowcolor{appendixboxbg}Own model & 75.00 & \textbf{90.00} & \textbf{52.25} & \textbf{66.78} \\
doubao-seed & \textbf{75.33} & 89.00 & 48.96 & 64.71 \\
\bottomrule
\end{tabular}
\end{table}

%% file: sections/AppendixC_Proof.tex
\section{Proof of Proposition 1}
\label{app:proof}

\paragraph{Proposition 1 (Restated).}
\textit{Under the grounding constraint of $G_0$ (Eq.~2), the G-ReAct reasoning state $S_i=(F_{\leq i},\mathcal{C}_i,\Sigma_i)$ is monotonically non-decreasing. For any $i<j$: (a) $F_i\subseteq F_j$; (b) $\gamma_c^{(i)}\leq\gamma_c^{(j)}$ for every candidate $c$ present in both rounds; and (c) any high-confidence candidate present in round $i$ is preserved in all subsequent rounds. As the verifiable facts are finite and $\gamma_c$ ranges over a bounded set, the state converges to a fixed point.}

\medskip
\noindent\textit{Proof.}
We establish the three properties in turn; the convergence claim then follows from boundedness and monotonicity.

\paragraph{Property (a).}
The append-only fact accumulation policy (Eq.~7) updates $F$ by a set union,
$F_{i+1}=F_i\cup F_i^{(\mathrm{new})}$, and never removes an existing fact. Hence $F_i\subseteq F_{i+1}$ for every $i$, and by induction $F_i\subseteq F_j$ whenever $i<j$. Since the verifiable facts for a given question are finite (tool responses carry bounded information), $\{F_i\}$ is a monotonically non-decreasing sequence bounded above by the finite set of all verifiable facts, and therefore converges to a fixed point.

\paragraph{Property (b).}
Hierarchical candidate merging (Eq.~8) sets $\gamma_m^{(i+1)}=\max\!\big(\gamma_m^{(i)},\gamma_m^{(\mathrm{new})}\big)$, so $\gamma_c^{(i)}\leq\gamma_c^{(i+1)}$ for every recurring candidate $c$. By induction this gives $\gamma_c^{(i)}\leq\gamma_c^{(j)}$ for $i<j$. Since $\gamma_c\in\{\text{high},\text{medium},\text{low}\}$ is a finite (hence bounded) ordered set, the non-decreasing sequence $\{\gamma_c^{(i)}\}$ is bounded and converges.

\paragraph{Property (c).}
The high-confidence candidate protection rule (Eq.~9) re-injects any candidate $c$ with $\gamma_c^{(i)}=\text{high}$ that is missing in $\mathcal{C}_{i+1}$, i.e.\ $\mathcal{C}_{i+1}(v)\leftarrow\mathcal{C}_{i+1}(v)\cup\{c\}$. Together with Property~(b), which ensures the confidence of a re-injected candidate does not decrease, this guarantees that every high-confidence candidate present in round $i$ survives in all subsequent rounds, both as an element of $\mathcal{C}$ and at its attained confidence level.

\smallskip
Since each component of $S_i$ is, by the above, monotonically non-decreasing and bounded under any execution trace, the state $S_i$ converges to a fixed point. This completes the proof. \hfill$\square$

%% file: sections/AppendixD_Implementation.tex
\section{Implementation Details}
\label{app:implementation}

\subsection{Training Setup}
The base model is Qwen3-30B-A3B-Thinking-2507, a mixture-of-experts model with 3B activated parameters. We perform full-parameter supervised fine-tuning (i.e., no parameter-efficient adapters) without reinforcement learning, using the Megatron framework (Megatron-LM with Transformer Engine, BF16 mixed precision) on 2$\times$8 NVIDIA A100 80GB GPUs (16 devices in total). Training uses a MoE-aware 3D parallel layout (tensor parallelism TP$=$4, expert parallelism EP$=$2, pipeline parallelism PP$=$2, expert tensor parallelism ETP$=$1, and context parallelism CP$=$1, yielding a data-parallel degree of 2) with a maximum sequence length of 128k tokens and full activation recomputation (one layer per stage, uniformly) for memory efficiency. Optimization uses AdamW with a peak learning rate of $1\times10^{-5}$ cosine-annealed to $1\times10^{-7}$, 6 linear warmup steps, and zero weight decay. We train for 3 epochs ($\approx$178 optimizer steps) with an effective batch size of 32 (global batch size 16, data-parallel degree 2), fixing the random seed to 1{,}234 for reproducibility. Each training sample contains a question $q$, a query graph $G_0$, and interleaved reasoning steps, tool calls, and full uncompressed tool responses, with a tool-call timeout of 10\,s.

\subsection{Training Data Construction Pipeline}
We construct a high-quality SFT trajectory dataset through a strict multi-stage filtering pipeline. Each trajectory is a multi-turn deep-search interaction over the shared tools \texttt{search} and \texttt{visit}, structured on a fixed query graph $G_0$ (\S3.1) and following the standard \texttt{think} $\to$ \texttt{tool\_call} $\to$ \texttt{tool\_response} cycle, terminating in an \texttt{answer} tag.

Two trajectory types are collected, distinguished by the reasoning state carried by the first \texttt{user} turn. \emph{(1)~From-scratch (single-round) trajectories}: the first \texttt{user} turn carries an empty state $S_0=\emptyset$ (no progress check, no candidates, no verified facts), and the solve completes within a single 64k-token round. \emph{(2)~State-rendered (warm-start) trajectories}: the first \texttt{user} turn carries a pre-computed state $S_i$ (a \texttt{best\_candidate\_set} with \texttt{exists\_fully\_satisfying\_candidate\_set}, per-node candidates with confidence ratings, and a verified-fact pool $F_{\leq i}$), produced by an evidence-driven state update $\Psi$ (\S3.3) from a preceding exploration round that did not converge within a single round, so the agent continues from an accumulated state rather than re-exploring from zero.

In the 1{,}898-trajectory pool (sampled from OpenSeeker-v1) the two types account for 1{,}156 (60.9\%) and 742 (39.1\%). Both terminate with an answer: type-1 trajectories reach a satisfying answer within a single round, whereas type-2 trajectories either converge after state-conditioned continuation or, when all $K$ rounds are exhausted without convergence, emit a best-effort answer from the accumulated state per \S3.4. Each sample contains the question $q$, the graph $G_0$, and interleaved reasoning steps, tool calls, and full uncompressed tool responses.

The filtering pipeline repairs non-standard tags (\texttt{think}/\texttt{tool\_call}/\texttt{tool\_response}/\texttt{answer} delimiters), removes invalid content (error markers, empty tool responses, and consecutive duplicate queries that signal retrieval loops), and replaces malformed expressions. The final output is a standardized JSON array retaining only the \texttt{messages} field, with interleaved \texttt{user}/\texttt{assistant} turns in which each tool response is injected as a \texttt{user} message.

\subsection{Trajectory-Construction Ablation}
\label{app:trajectory_ablation}
To isolate the contribution of graph-structured reasoning, we compare two trajectory-construction methods on the same 618 randomly sampled DeepDive QA pairs under identical training and inference configurations: standard ReAct without graph guidance, and G-ReAct with explicit query-graph conditioning. As Table~\ref{tab:ablation_greact} shows, G-ReAct improves pass@1 by $+$13.5 on BrowseComp-ZH (45.33 vs.\ 31.83) and $+$7.5 on BrowseComp (28.46 vs.\ 20.95), with matching accuracy gains ($+$12.8 and $+$7.4). On XBench, G-ReAct attains the best pass@3 (\textbf{85.00} vs.\ 79.00) and accuracy (\textbf{67.67} vs.\ 66.00) despite a marginally lower pass@1 (70.00 vs.\ 73.00), indicating that graph guidance favors verifying consolidated candidates over early guessing. Graph guidance also reduces reasoning turns on BrowseComp-ZH (22.53 vs.\ 26.29) and XBench (\textbf{15.66} vs.\ 17.50), while taking more turns on BrowseComp (35.84 vs.\ 27.96) for a $+$7.4 accuracy gain via explicit intermediate-step verification.

\begin{table}[!ht]
\centering
\caption{Trajectory-construction ablation on 618 DeepDive QA pairs under identical training and inference configurations (64k context). Both methods use the same base model, SFT recipe, and evaluation protocol. Avg.\ Turns = average interaction turns for first-attempt correct solutions. $\uparrow$/$\downarrow$ = higher/lower is better. Best per benchmark and metric is \textbf{bolded}.}
\label{tab:ablation_greact}
\setlength{\tabcolsep}{2.8pt}
\renewcommand{\arraystretch}{1.1}
\footnotesize
\begin{tabular}{@{}llrrrr@{}}
\toprule
Benchmark & Method & p@1$\uparrow$ & p@3$\uparrow$ & Acc$\uparrow$ & Turn$\downarrow$ \\
\midrule
\multirow{2}{*}{BrowseComp-ZH} & Standard ReAct     & 31.83 & 48.10 & 31.26 & 26.29 \\
                               & \cellcolor{appendixboxbg}\textbf{G-ReAct (Ours)} & \cellcolor{appendixboxbg}\textbf{45.33} & \cellcolor{appendixboxbg}\textbf{61.94} & \cellcolor{appendixboxbg}\textbf{44.06} & \cellcolor{appendixboxbg}\textbf{22.53} \\
\midrule
\multirow{2}{*}{BrowseComp}    & Standard ReAct     & 20.95 & 32.41 & 19.76 & \textbf{27.96} \\
                               & \cellcolor{appendixboxbg}\textbf{G-ReAct (Ours)} & \cellcolor{appendixboxbg}\textbf{28.46} & \cellcolor{appendixboxbg}\textbf{39.53} & \cellcolor{appendixboxbg}\textbf{27.14} & \cellcolor{appendixboxbg}35.84 \\
\midrule
\multirow{2}{*}{XBench}        & Standard ReAct     & \textbf{73.00} & 79.00 & 66.00 & 17.50 \\
                               & \cellcolor{appendixboxbg}\textbf{G-ReAct (Ours)} & \cellcolor{appendixboxbg}70.00 & \cellcolor{appendixboxbg}\textbf{85.00} & \cellcolor{appendixboxbg}\textbf{67.67} & \cellcolor{appendixboxbg}\textbf{15.66} \\
\bottomrule
\end{tabular}
\end{table}

\subsection{Computational Budgets}
Per-question limits: at most $K = 5$ refinement rounds; at most $N_{\max} = 100$ LLM calls per round; a per-round timeout of $T_{\max} = 20$ minutes; and a per-round context window of 128k tokens. A state update is triggered when the context reaches $L_{\max} = 64\text{k}$ tokens (50\% of the window, leaving headroom for the update prompt), when an exploration round explicitly terminates, or when the tool-call budget is exhausted.

\subsection{Tool Specifications}
\textbf{Search} performs batched Google web search, supports multiple parallel queries, returns the top-10 results per query, and deduplicates identical queries via a TTL cache. \textbf{Visit} processes a URL through the pipeline \texttt{crawler} $\to$ \texttt{Jina} $\to$ \texttt{LLM summary} $\to$ \texttt{JSON output}, yielding a \texttt{rational}/\texttt{evidence}/\texttt{summary} JSON object. To break retrieval loops, the system snapshots the most recent tool call and, upon detecting two consecutive identical calls, returns a nudge that prompts the model toward new search directions.


%% file: sections/AppendixE_Benchmarks.tex
\section{Benchmark Details}
\label{app:benchmarks}

\textbf{BrowseComp} and \textbf{BrowseComp-ZH} are open-ended web-browsing benchmarks (in English and Chinese, respectively) that require multi-hop retrieval, cross-document reasoning, and fact verification. Questions are multi-hop in nature and involve multiple unknown entities, implicit relations, and ambiguous constraints, which the agent must resolve through several rounds of search, verification, and chaining.

\textbf{GAIA} is a general AI-assistant benchmark requiring heterogeneous tool use and multi-step compositional reasoning. We adopt its 103-sample text-only validation subset; all other benchmarks use their complete test sets.

\textbf{XBench-DeepSearch} (XBench-DS) is a comprehensive information-synthesis benchmark requiring agents to aggregate and reconcile evidence from multiple, cross-domain sources.

%% file: sections/AppendixF_Schema.tex
\section{Query Graph Schema}
\label{app:schema}
This appendix details the query-graph schema introduced in the main text (\S3.1): the JSON structure, the full $21$-type relation ontology, and the faithfulness constraints that every $G_0$ must satisfy.

\subsection{Graph Structure Definition}
The query graph $G_0 = (V_0, E_0)$ follows a strict JSON schema designed for faithful constraint representation:

\begin{tcblisting}{listing only,
  colback=appendixboxbg, colframe=appendixboxframe,
  boxrule=.4pt, arc=2pt, boxsep=0pt,
  left=6pt, right=6pt, top=2pt, bottom=2pt,
  title=Query Graph JSON Schema,
  fonttitle=\bfseries\small,
  lefttitle=6pt, righttitle=6pt, toptitle=2pt, bottomtitle=2pt,
  coltitle=white, colbacktitle=appendixboxtitle,
  listing options={style=appendixplaincode, basicstyle=\scriptsize\ttfamily,
    breaklines=true, breakatwhitespace=true, breakindent=0pt,
    showstringspaces=false, columns=fullflexible, keepspaces=true}}
{
  "nodes": [
    {
      "id": "node name",
      "type": "abstract | entity",
      "constraints": "must come from original text snippets, may be empty string",
      "anchor_text": "must be a contiguous substring of the original text"
    }
  ],
  "edges": [
    {
      "id": "sequence number",
      "source": "node1",
      "target": "node2",
      "relation": "relation type",
      "value": "must come from original text snippets, may be empty string",
      "anchor_text": "must be a contiguous substring of the original text"
    }
  ]
}
\end{tcblisting}

\subsection{Relation Ontology (21 Types)}
The relation set $\mathcal{R}=\{r_1,\ldots,r_{21}\}$ sketched in the main text (\S3.1) is fully specified here. It is organized into six semantic families, with \texttt{associated\_with} as a fallback for weakly-typed edges:
\begingroup
\renewcommand{\_}{\textunderscore\discretionary{}{}{}}
\newcommand{\rt}[1]{\texttt{#1}}
\begin{itemize}\raggedright\setlength{\itemsep}{3pt}\setlength{\parskip}{0pt}\setlength{\labelsep}{4pt}
    \item \textbf{Compositional:} \rt{has\_part}, \rt{has\_member}, \rt{has\_author}, \rt{has\_reference}, \rt{has\_character}, \rt{has\_protagonist}.
    \item \textbf{Attributional:} \rt{published\_in}, \rt{reference\_published\_in}, \rt{affiliated\_with}.
    \item \textbf{Usage:} \rt{uses}, \rt{uses\_engine}, \rt{implemented\_in}.
    \item \textbf{Media:} \rt{appears\_in}, \rt{based\_on}, \rt{named\_after}, \rt{related\_to}.
    \item \textbf{Constraint:} \rt{has\_property}, \rt{quantity\_constraint}, \rt{time\_constraint}, \rt{index\_constraint}.
    \item \textbf{Fallback:} \rt{associated\_with} (only when no explicit semantic relation applies).
\end{itemize}
\endgroup
Usage rules: (1) prefer semantically explicit relations; (2) use \texttt{associated\_with} only when no explicit match exists, in which case the original-text snippet must be preserved in \texttt{edge.value}.

\subsection{Faithfulness Constraints}
As stated by the grounding principle in the main text (\S3.1, Eq.~1), every structural element must satisfy:
\begin{equation*}
\forall x \in V_0 \cup E_0,\quad \mathrm{anchor}(x) \subseteq \mathrm{span}(q).
\end{equation*}

\noindent\textbf{Prohibited Behaviors.}
\begin{itemize}
    \item Introducing information not present in the original text;
    \item Synonym rewriting, normalization, or translation (e.g., ``WSDM2023'' must not be rewritten as ``WSDM 2023''; ``HKU'' must not be expanded unless the original text states the full name);
    \item Adding extra nodes for ``completeness'';
    \item Inferring quantities from common sense (e.g., ``half = 3'') unless the original text provides the number.
\end{itemize}
These constraints ensure $G_0$ is a strictly grounded structural projection of the input question rather than an injection of external knowledge.

%% file: sections/AppendixG_Prompts.tex
\section{Prompt Templates}
\label{app:prompts}

\subsection{System Prompt}
\begin{tcolorbox}[colback=appendixboxbg, colframe=appendixboxframe, boxrule=0.4pt, arc=2pt, left=6pt, right=6pt, top=4pt, bottom=4pt, fonttitle=\bfseries\small, title=System Prompt, title after break={}, coltitle=white, colbacktitle=appendixboxtitle, breakable]
\small\sloppy
You are a rea\-son\-ing expert tasked with solv\-ing com\-plex prob\-lems through struc\-tured think\-ing and ev\-i\-dence-based answers.

\medskip
Begin by out\-lin\-ing your rea\-son\-ing pro\-cess within \texttt{<think>} tags. In this sec\-tion, explain your cur\-rent under\-stand\-ing, what in\-for\-ma\-tion is miss\-ing, and what needs to be ver\-i\-fied.

\medskip
When ex\-ter\-nal in\-for\-ma\-tion is re\-quired, spec\-i\-fy a tool in\-vo\-ca\-tion inside \texttt{<tool\_call>} tags, de\-scrib\-ing the func\-tion name and argu\-ments in JSON format.

\medskip
After the en\-vi\-ron\-ment in\-jects a \texttt{<tool\_re\-sponse>}, start a new \texttt{<think>} block that in\-te\-grates that re\-sponse and up\-dates your plan; you may then is\-sue an\-oth\-er \texttt{<tool\_call>} if needed. Repeat this strict cy\-cle of \texttt{<think>} $\rightarrow$ \texttt{<tool\_call>} $\rightarrow$ wait-for-\texttt{<tool\_re\-sponse>} $\rightarrow$ \texttt{<think>} until you have ga\-thered suf\-fi\-cient ev\-i\-dence to reach a con\-fi\-dent con\-clu\-sion.

\medskip
When your rea\-son\-ing is com\-plete, pro\-duce the final answer en\-closed in \texttt{<answer>}\texttt{</answer>} tags. The final answer must be con\-cise, log\-i\-cal\-ly con\-sis\-tent, and di\-rect\-ly sup\-port\-ed by the ga\-thered in\-for\-ma\-tion.

\medskip
\noindent\textbf{Tools.} You may call one or more functions to assist with the user queries. You are pro\-vid\-ed with func\-tion sig\-na\-tures within \texttt{<tools>}\texttt{</tools>} XML tags:
\begin{tcblisting}{listing only, colback=appendixboxbg, colframe=appendixboxbg, boxrule=0pt, arc=0pt, left=0pt, right=0pt, top=0pt, bottom=0pt, breakable, listing options={style=appendixplaincode, basicstyle=\scriptsize\ttfamily, breaklines=true, breakatwhitespace=true, breakindent=0pt, aboveskip=0pt, belowskip=0pt, showstringspaces=false, columns=fullflexible, keepspaces=true}}
<tools>
{"type": "function", "function": {
  "name": "search",
  "description": "Perform Google web searches then returns a string of the top search results. Accepts multiple queries.",
  "parameters": {
    "type": "object",
    "properties": {
      "query": {
        "type": "array",
        "items": {"type": "string", "description": "The search query."},
        "minItems": 1,
        "description": "The list of search queries."
      }
    },
    "required": ["query"]
  }
}}
{"type": "function", "function": {
  "name": "visit",
  "description": "Visit web page(s) and return the summary of the content.",
  "parameters": {
    "type": "object",
    "properties": {
      "url": {
        "type": "array",
        "items": {"type": "string"},
        "description": "The URL(s) of the web page(s) to visit. Can be a single URL or an array of URLs."
      },
      "goal": {
        "type": "string",
        "description": "The specific information goal for visiting web page(s)."
      }
    },
    "required": ["url", "goal"]
  }
}}
</tools>
\end{tcblisting}
\noindent For each function call, return a JSON object with function name and arguments within \texttt{<tool\_call>}\texttt{</tool\_call>} XML tags:
\begin{tcblisting}{listing only, colback=appendixcodebg, colframe=appendixcodebg, boxrule=0pt, arc=0pt, left=0pt, right=0pt, top=0pt, bottom=0pt, breakable, listing options={style=appendixcode, basicstyle=\scriptsize\ttfamily, breaklines=true, breakatwhitespace=true, breakindent=0pt, aboveskip=0pt, belowskip=0pt, showstringspaces=false, columns=fullflexible, keepspaces=true}}
{"name": <function-name>, "arguments": <args-json-object>}
\end{tcblisting}
\end{tcolorbox}

\subsection{G-ReAct Context Template}
\begin{tcolorbox}[colback=appendixboxbg, colframe=appendixboxframe, boxrule=0.4pt, arc=2pt, left=6pt, right=6pt, top=4pt, bottom=4pt, title=G-ReAct Context Template, title after break={}, fonttitle=\bfseries\small, coltitle=white, colbacktitle=appendixboxtitle, breakable]
\small\sloppy
\textbf{[Question]}\par\nobreak\smallskip
\texttt{\{question\}}

\bigskip
\textbf{[Base Query Graph]}\par\nobreak\smallskip
This graph is extracted from the question. It is NOT evidence. Use it as a constraint scaffold for slot filling.\par\nobreak\smallskip
\texttt{\{query\_graph\_json\}}

\bigskip
\textbf{[Progress Check: Candidate Sets \& Missing Constraints]}\par\nobreak\smallskip
This section shows a HYPOTHESIS based on evidence so far. It may be wrong. Use it as reference, not as ground truth. If your search reveals contradictions, update your reasoning.\par\nobreak\smallskip
\texttt{\{solution\_check\_summary\}}

\bigskip
\textbf{[Node Candidates (Slot Filling State)]}\par\nobreak\smallskip
Candidates are grouped by \texttt{node\_id}. These are hypotheses --- they may be incomplete or wrong. If you find a better candidate not listed here, pursue it.\par\nobreak\smallskip
\texttt{\{node\_candidates\_lines\}}

\bigskip
\textbf{[Verified Facts (Evidence Pool)]}\par\nobreak\smallskip
All facts below are explicitly verified by \texttt{tool\_response}. Do NOT repeat searches for facts already verified.\par\nobreak\smallskip
\texttt{\{verified\_facts\_lines\}}

\bigskip
\textbf{[Instruction]} You must solve the question using evidence-based reasoning and tool calls (\texttt{search}/\texttt{visit}). \texttt{tool\_response} is the only valid evidence source. Do NOT restart from scratch.\par\nobreak\smallskip
\textbf{Suggested workflow:}
\begin{enumerate}\setlength{\itemsep}{2pt}\setlength{\parskip}{0pt}
    \item \textbf{Sanity check.} Before investing effort, quickly assess the \texttt{best\_}\discretionary{}{}{}\texttt{candidate\_}\discretionary{}{}{}\texttt{set} above:
    \begin{itemize}\setlength{\itemsep}{1pt}\setlength{\parsep}{0pt}\setlength{\topsep}{0pt}\setlength{\partopsep}{0pt}
        \item Does it make intuitive sense given the question?
        \item Are there $2$ or more \texttt{missing\_constraints} with confidence $\leq$ medium?
        \item If it seems wrong or has too many missing constraints, do \textbf{not} blindly verify it. Instead, search broadly for alternative candidates first.
    \end{itemize}
    \item If \texttt{best\_candidate\_set} confidence is \texttt{HIGH}: prioritize verifying its \texttt{missing\_constraints} with targeted searches.
    \item If confidence is \texttt{MEDIUM}: spend $1$--$2$ searches verifying the candidate. If verification fails or contradicts, immediately switch to broad exploration.
    \item If confidence is \texttt{LOW} or candidates seem wrong: ignore current candidates and search broadly for alternative directions.
    \item You are \textbf{not} bound by the \texttt{candidate\_sets} --- they are hypotheses, not conclusions. If new evidence contradicts them, pursue the new direction immediately.
    \item \textbf{Balance your evidence.} If verified facts are heavily biased toward certain candidates while others have zero supporting facts, actively search for facts about those underrepresented candidates.
    \item Stop once you have enough evidence to answer confidently.
    \item When providing the final answer, include the answer in the language that matches the question. If the question is in Chinese, provide the Chinese form of names/terms; if an entity has well-known Chinese translations, include them.
\end{enumerate}
\end{tcolorbox}

\subsection{Query Graph Construction Prompt}
The initializer~$\phi$ is realized by a single prompt that casts the LLM as a ``Query Constraint Graph Builder.'' The full prompt is given below.
\begin{tcolorbox}[colback=appendixboxbg, colframe=appendixboxframe, boxrule=0.4pt, arc=2pt, left=6pt, right=6pt, top=4pt, bottom=4pt, title=Query Graph Construction Prompt, title after break={}, fonttitle=\bfseries\small, coltitle=white, colbacktitle=appendixboxtitle, breakable]
\small\sloppy
You are a Query Constraint Graph Builder.

\medskip
Your task is: based solely on the input question text, construct the ``Constraint Structure Graph (Query Graph)'' for that question.\par\nobreak\smallskip
The Query Graph only represents the logical constraint structure of the question. It does not perform any reasoning, does not introduce external knowledge, and does not generate answer candidates.

\noindent\textbf{[Highest Priority Rule: Must Be Traceable (Grounded Extraction Only)]}
\medskip
\noindent Every node, every edge, and every constraint you output must have a directly corresponding ``anchor text (\texttt{anchor\_text})'' findable in the original question.\par\nobreak\smallskip
If any piece of information cannot be mapped to a corresponding snippet in the original text, it is forbidden to output that information.

\medskip
\textbf{[Prohibited Behaviors.]}
\begin{enumerate}\setlength{\itemsep}{1pt}\setlength{\parskip}{0pt}\setlength{\topsep}{0pt}
    \item Introducing any information not present in the original text.
    \item Synonym rewriting, normalization, translation, or completion (e.g., ``WSDM2023'' must not be rewritten as ``WSDM 2023''; ``HKU'' must not be replaced with ``The University of Hong Kong'' unless the original text explicitly states it).
    \item Adding extra nodes to make the graph complete.
    \item Inferring quantities from common sense (e.g., ``half = 3''), unless the original text explicitly provides the number.
\end{enumerate}

\medskip
\textbf{[Core Objectives.]}
\begin{enumerate}\setlength{\itemsep}{1pt}\setlength{\parskip}{0pt}\setlength{\topsep}{0pt}
    \item Transform the question into a structured constraint graph.
    \item The \texttt{Answer} node must exist.
    \item Cover all constraints in the original question text.
    \item The graph structure should be as compact as possible without omitting constraints.
    \item Each node/edge must carry an \texttt{anchor\_text} (original-text snippet).
\end{enumerate}

\medskip
\textbf{[Two-Phase Construction Process (Must Execute)]}
\medskip
\noindent\textbf{Step 1: Constraint Span Extraction (Internal step, not output)}
\begin{itemize}\setlength{\itemsep}{1pt}\setlength{\parskip}{0pt}\setlength{\topsep}{0pt}
    \item Extract all constraint phrases from the original question text (verbatim copy).
    \item These phrases include but are not limited to: time, location, quantity, author order, identity, attribute requirements, comparative relationships, etc.
    \item Only phrases that appear in the original text may be extracted.
\end{itemize}

\noindent\textbf{Step 2: Graph Construction (Output Query Graph)}
\begin{itemize}\setlength{\itemsep}{1pt}\setlength{\parskip}{0pt}\setlength{\topsep}{0pt}
    \item Only phrases extracted in Step 1 may be used to construct nodes/edges.
    \item Each node/edge must provide an \texttt{anchor\_text} to prove that the structure comes from the original text.
    \item Reference sub-steps:
    \begin{itemize}\setlength{\itemsep}{1pt}\setlength{\parskip}{0pt}\setlength{\topsep}{0pt}
        \item \textbf{Step 2.1:} Create the \texttt{Answer} node (abstract)
        \item \textbf{Step 2.2:} Identify explicit entities in constraint snippets, create entity nodes
        \item \textbf{Step 2.3:} Identify key logical roles implied in constraint snippets, create abstract nodes as needed (keep minimal but complete)
        \item \textbf{Step 2.4:} Identify structural relationships, connect nodes with edges, prefer explicit relations
        \item \textbf{Step 2.5:} Write quantity, index, time constraints into \texttt{node.constraints} or \texttt{edge.value} to avoid creating redundant nodes
        \item \textbf{Step 2.6:} Check whether compound relationships have not been split; if so, they must be split into multiple edges
    \end{itemize}
\end{itemize}

\textbf{[Node Field Definitions]}

\noindent \texttt{node.type} $\in$ \{ ``abstract'', ``entity'' \}

\smallskip
\noindent\textbf{1. abstract node}
\par\nobreak\smallskip
\noindent\emph{Definition:} Represents logical variables, reasoning slots, or structural roles in the question; does not correspond to specific real-world entities.
\par\medskip
\noindent\texttt{node.id:}
\begin{itemize}\setlength{\itemsep}{1pt}\setlength{\parskip}{0pt}\setlength{\topsep}{0pt}
    \item The Answer node id must be fixed as ``Answer''.
    \item Other abstract nodes must be semantically clear and stable (\texttt{Paper}, \texttt{Author\_1}, \texttt{Person}, \texttt{Festival\_1}, etc.).
    \item Indexed objects must be explicitly numbered (\texttt{Author\_4} represents ``the fourth author'').
    \item \texttt{Node1} / \texttt{Obj} / \texttt{Item} and other meaningless names are forbidden.
\end{itemize}
\noindent\texttt{node.constraints:}
\begin{itemize}\setlength{\itemsep}{1pt}\setlength{\parskip}{0pt}\setlength{\topsep}{0pt}
    \item Must be excerpts or combinations from the original text snippets (rewriting is forbidden).
    \item Only describe the node's own attributes.
    \item If there are no constraints, use an empty string ``''.
\end{itemize}
\noindent\texttt{node.anchor\_text:}
\begin{itemize}\setlength{\itemsep}{1pt}\setlength{\parskip}{0pt}\setlength{\topsep}{0pt}
    \item Must be a contiguous substring of the original question text (substring match).
    \item Used to prove that the node exists or that the node's constraints come from the original text.
    \item If the node is a purely structural node (e.g., \texttt{Paper}), \texttt{anchor\_text} must still be found from a snippet in the original text that supports its existence (e.g., ``a certain paper'').
\end{itemize}

\smallskip
\noindent\textbf{2. entity node}
\par\nobreak\smallskip
\noindent\emph{Definition:} Specific named entities explicitly appearing in the question (conference names, school names, person names, etc.).
\par\medskip
\noindent\texttt{node.id:}
\begin{itemize}\setlength{\itemsep}{1pt}\setlength{\parskip}{0pt}\setlength{\topsep}{0pt}
    \item Must be exactly identical to the original question text (substring match).
    \item Any form of rewriting, translation, or normalization is forbidden.
\end{itemize}
\noindent\texttt{node.constraints:}
\begin{itemize}\setlength{\itemsep}{1pt}\setlength{\parskip}{0pt}\setlength{\topsep}{0pt}
    \item May be an empty string ``''.
    \item Or modifier information adjacent to the entity in the original text (must come from the original text).
\end{itemize}
\noindent\texttt{node.anchor\_text:}
\begin{itemize}\setlength{\itemsep}{1pt}\setlength{\parskip}{0pt}\setlength{\topsep}{0pt}
    \item Must equal \texttt{node.id} (i.e., the entity string appearing in the original text).
\end{itemize}

\textbf{[Relation Field Definitions]}

\smallskip
\noindent\texttt{edge.id:} Must be unique and non-repeating, incrementing from 0.

\smallskip
\noindent\texttt{edge.source} / \texttt{edge.target} must exist in nodes.

\smallskip
\noindent\texttt{edge.relation} $\in$ \{``has\_\penalty0part'', ``has\_\penalty0member'', ``has\_\penalty0author'', ``has\_\penalty0reference'', ``has\_\penalty0character'', ``has\_\penalty0protagonist'', ``published\_\penalty0in'', ``reference\_\penalty0published\_\penalty0in'', ``affiliated\_\penalty0with'', ``uses'', ``uses\_\penalty0engine'', ``implemented\_\penalty0in'', ``appears\_\penalty0in'', ``based\_\penalty0on'', ``named\_\penalty0after'', ``related\_\penalty0to'', ``has\_\penalty0property'', ``quantity\_\penalty0constraint'', ``time\_\penalty0constraint'', ``index\_\penalty0constraint'', ``associated\_\penalty0with''\} (the $21$-type ontology in \S B.2).

\smallskip
\noindent Relation usage rules:
\begin{enumerate}\setlength{\itemsep}{1pt}\setlength{\parskip}{0pt}\setlength{\topsep}{0pt}
    \item Prefer semantically explicit relations.
    \item Only use \texttt{associated\_with} when no explicit semantic relation can be matched.
    \item When using \texttt{associated\_with}, the original text description snippet must be preserved in \texttt{edge.value} (directly copy from the original text).
\end{enumerate}

\noindent\texttt{edge.value:}
\begin{itemize}\setlength{\itemsep}{1pt}\setlength{\parskip}{0pt}\setlength{\topsep}{0pt}
    \item Must come from the original question text snippets (verbatim copy or excerpt, rewriting is forbidden).
    \item If there is no supplementary description, use an empty string ``''.
\end{itemize}
\noindent\texttt{edge.anchor\_text:}
\begin{itemize}\setlength{\itemsep}{1pt}\setlength{\parskip}{0pt}\setlength{\topsep}{0pt}
    \item Must be a contiguous substring of the original question text (substring match), used to prove that the edge exists.
\end{itemize}

\textbf{[Key Modeling Rules (Must Follow)]}
\begin{enumerate}\setlength{\itemsep}{2pt}\setlength{\parskip}{0pt}\setlength{\topsep}{0pt}
    \item \emph{Index constraints must be explicitly modeled.}
    \begin{itemize}\setlength{\itemsep}{1pt}\setlength{\parskip}{0pt}\setlength{\topsep}{0pt}
        \item If the original text contains descriptions like ``the $N$-th author / the fourth reference'', and it is a necessary intermediate node for reasoning, the corresponding abstract node \texttt{Author\_N} / \texttt{Reference\_N} must be created.
        \item The original index phrase must be preserved in \texttt{node.constraints} or \texttt{edge.value}.
    \end{itemize}
    \item \emph{Abstract nodes are allowed, but must have a corresponding \texttt{anchor\_text}.}
    \begin{itemize}\setlength{\itemsep}{1pt}\setlength{\parskip}{0pt}\setlength{\topsep}{0pt}
        \item If \texttt{anchor\_text} cannot be found in the original text, creating that node is forbidden.
    \end{itemize}
\end{enumerate}

\textbf{[Output Format (Strict JSON, Output JSON Only)]}
\begin{tcblisting}{listing only, colback=appendixcodebg, colframe=appendixcodebg, boxrule=0pt, arc=0pt, left=0pt, right=0pt, top=0pt, bottom=0pt, breakable, listing options={style=appendixcode, basicstyle=\scriptsize\ttfamily, breaklines=true, breakatwhitespace=true, breakindent=0pt, aboveskip=0pt, belowskip=0pt, showstringspaces=false, columns=fullflexible, keepspaces=true}}
{
  "nodes": [
    {
      "id": "node name",
      "type": "abstract | entity",
      "constraints": "must come from original text snippets, may be empty string",
      "anchor_text": "must be a contiguous substring of the original text"
    }
  ],
  "edges": [
    {
\end{tcblisting}
\begin{tcblisting}{listing only, colback=appendixcodebg, colframe=appendixcodebg, boxrule=0pt, arc=0pt, left=0pt, right=0pt, top=0pt, bottom=0pt, breakable, listing options={style=appendixcode, basicstyle=\scriptsize\ttfamily, breaklines=true, breakatwhitespace=true, breakindent=0pt, aboveskip=0pt, belowskip=0pt, showstringspaces=false, columns=fullflexible, keepspaces=true}}
      "id": "sequence number",
      "source": "node1",
      "target": "node2",
      "relation": "relation type",
      "value": "must come from original text snippets, may be empty string",
      "anchor_text": "must be a contiguous substring of the original text"
    }
  ]
}
\end{tcblisting}
\end{tcolorbox}

\subsection{State Transition Prompt}
The state-transition operator~$\Psi$ is realized by a prompt that casts the LLM as a strict ``Search Trace Fact Extraction and Graph Evolution Analysis Assistant.'' Its full content is given below.
\begin{tcolorbox}[colback=appendixboxbg, colframe=appendixboxframe, boxrule=0.4pt, arc=2pt, left=6pt, right=6pt, top=4pt, bottom=4pt, title=State Transition Prompt, title after break={}, fonttitle=\bfseries\small, coltitle=white, colbacktitle=appendixboxtitle, breakable]
\small\sloppy
You are a strict ``Search Trace Fact Extraction and Graph Evolution Analysis Assistant.'' Your task is: extract explicit facts from multi-turn search/tool-call traces, and perform structural alignment and next-step graph-evolution analysis based on the Query Graph. You must strictly follow the four steps below.

\medskip
\textbf{[Input.]}
\begin{itemize}\setlength{\itemsep}{1pt}\setlength{\parskip}{0pt}\setlength{\topsep}{0pt}
    \item \emph{Query Graph:} nodes (abstract / entity), edges.
    \item \emph{Search Trace:} \texttt{assistant\_\penalty0think}, \texttt{tool\_\penalty0call}, \texttt{tool\_\penalty0response}.
\end{itemize}

\medskip
\textbf{[Global Principles.]}
\begin{enumerate}\setlength{\itemsep}{1pt}\setlength{\parskip}{0pt}\setlength{\topsep}{0pt}
    \item \texttt{tool\_\penalty0response} is the sole source of facts.
    \item \texttt{think} may only be used to determine which \texttt{tool\_\penalty0response} information is relevant to solving the problem (relevance filtering).
    \item All facts must not involve reasoning, summarization, or expansion; they must come from a single \texttt{tool\_\penalty0response}.
    \item Modifying the graph's semantic structure is not allowed (only priority modifications and conservative evolution).
    \item Inventing new nodes / edges is not allowed.
    \item Any ``binding relationship'' between candidate entities must come from explicit fact evidence; subjective splicing is not allowed.
\end{enumerate}

\medskip
\textbf{[Step 1: Fact Extraction (must be done first).]} Extract all atomic facts from \texttt{tool\_\penalty0response}.
\emph{Requirements:} single semantic unit; cannot be further decomposed; must not contain causation / evaluation / summarization / reasoning; must have a directly supporting sentence findable in \texttt{tool\_\penalty0response}; should be as close to the original wording as possible (minor compression allowed without changing semantics).
\emph{Forbidden:} summary sentences; causal sentences (e.g., ``led to'', ``therefore''); evaluative sentences (e.g., ``important'', ``significant''); multi-fact merged sentences.
\emph{Output:} \texttt{verified\_\penalty0facts}. Each fact must contain \texttt{fact} (atomic fact) and \texttt{evidence\_\penalty0quote} (verbatim quote from \texttt{tool\_\penalty0response}).

\medskip
\textbf{[Step 2: Candidate Entity Aggregation.]} Based on \texttt{verified\_\penalty0facts}, align with the Query Graph and categorize information into candidate entities. \emph{Output:} \texttt{node\_\penalty0candidates}. Each node contains \texttt{node\_\penalty0id} and a \texttt{candidates} list; each candidate contains \texttt{name}, \texttt{supporting\_\penalty0constraints}, \texttt{violating\_\penalty0constraints}, and \texttt{confidence} (\texttt{high} / \texttt{medium} / \texttt{low}).
\emph{Rules:}
\begin{enumerate}\setlength{\itemsep}{1pt}\setlength{\parskip}{0pt}\setlength{\topsep}{0pt}
    \item \texttt{node\_\penalty0id} must come from \texttt{query\_graph.nodes[].id}.
    \item \texttt{candidate.name} must come from entities/values explicitly appearing in \texttt{tool\_\penalty0response}; inference-based completion is not allowed.
    \item Merging different entities is not allowed (unless \texttt{tool\_\penalty0response} explicitly states they are the same object).
    \item \texttt{supporting\_\penalty0constraints} / \texttt{violating\_\penalty0constraints} must be verbatim from \texttt{query\_graph.nodes[].constraints} or \texttt{query\_graph.edges[].value}.
    \item The determination of supporting / violating constraints must be supported by \texttt{verified\_\penalty0facts} evidence; \texttt{evidence\_\penalty0facts} must be exactly equal to a fact in \texttt{verified\_\penalty0facts} (verbatim match).
    \item \emph{Confidence rules:} \texttt{high} --- clear evidence supports that the candidate is strongly related to core constraints, with no conflicts; \texttt{medium} --- partial evidence support, but key information is still missing; \texttt{low} --- evidence is very weak, too much is missing, or conflicts exist.
\end{enumerate}

\medskip
\textbf{[Step 3: Global Consistency Check (Solution Check / Candidate Sets).]} Based on the Query Graph, \texttt{verified\_\penalty0facts}, and \texttt{node\_\penalty0candidates}, determine whether a candidate set exists that can satisfy all constraints of the graph.
\emph{You must output:} (A) \texttt{global\_\penalty0missing\_\penalty0constraints} --- constraints still missing globally; (B) \texttt{candidate\_\penalty0sets} --- multiple possible cross-node combination candidate sets; (C) whether a fully satisfying candidate set exists.
\par\nobreak\smallskip
\emph{Candidate-set definition.} A \texttt{candidate\_\penalty0set} is a cross-node ``binding hypothesis combination,'' e.g., a \texttt{mapping} from \texttt{Singer} to ``Liang Long'' and \texttt{Representative\_\penalty0Work\_\penalty01} to a specific work.
\emph{Candidate-set rules:}
\begin{enumerate}\setlength{\itemsep}{1pt}\setlength{\parskip}{0pt}\setlength{\topsep}{0pt}
    \item The keys of \texttt{mapping} must come from \texttt{query\_graph.nodes[].id}.
    \item The values of \texttt{mapping} must come from \texttt{candidate.name} already appearing in \texttt{node\_\penalty0candidates}.
    \item Arbitrary splicing of \texttt{mapping} is not allowed: if there is no explicit fact proving that candidates of two nodes belong to the same object chain, they must not be placed in the same \texttt{candidate\_\penalty0set}.
    \item \texttt{supported\_\penalty0constraints} / \texttt{missing\_\penalty0constraints} / \texttt{violating\_\penalty0constraints} must strictly reference \texttt{query\_graph.nodes[].constraints} or \texttt{query\_graph.edges[].value} (verbatim snippets).
    \item A \texttt{candidate\_\penalty0set} can only judge support/\mbox{conflict}/missing based on \texttt{verified\_\penalty0facts}; speculating ``possibly satisfies'' is not allowed.
    \item If \texttt{violating\_\penalty0constraints} exist and the conflicting constraint is a core constraint, confidence must be \texttt{low}; if it is a non-core auxiliary constraint, confidence may be \texttt{medium}.
    \item \texttt{candidate\_\penalty0set} confidence $\in$ \{\texttt{high}, \texttt{medium}, \texttt{low}\}: \texttt{high} --- most key constraints are supported, few missing, no conflicts; \texttt{medium} --- key missing constraints exist, but partial strong evidence supports the chain; \texttt{low} --- too many missing constraints, conflicts exist, or the evidence chain is very weak.
\end{enumerate}

\medskip
\textbf{[Step 4: Graph Simplification (default output \texttt{null}).]} You may output a \texttt{graph\_\penalty0simplification} field for ``conservative pruning suggestions,'' but it must satisfy: lightweight pruning (only delete redundant nodes/edges when obviously redundant) and lightweight merging (only merge nodes/edges when semantically equivalent). Graph simplification must be very cautious --- if redundancy cannot be confirmed, \texttt{null} must be output.

\medskip
\textbf{[Output Content.]} The following four parts must be output: (1) \texttt{verified\_\penalty0facts}; (2) \texttt{node\_\penalty0candidates}; (3) \texttt{solution\_\penalty0check} (containing \texttt{exists\_\penalty0fully\_\penalty0satisfying\_\penalty0candidate\_\penalty0set}, \texttt{global\_\penalty0missing\_\penalty0constraints}, \texttt{candidate\_\penalty0sets}, \texttt{best\_\penalty0candidate\_\penalty0set}, \texttt{reason}); (4) \texttt{graph\_\penalty0simplification}.

\medskip
\textbf{[Strict Constraints (reiterated).]}
\begin{enumerate}\setlength{\itemsep}{1pt}\setlength{\parskip}{0pt}\setlength{\topsep}{0pt}
    \item Only JSON may be output; explanatory text is not allowed.
    \item Reasoning is not allowed; completing missing information is not allowed.
    \item \texttt{verified\_\penalty0facts} must all come from \texttt{tool\_\penalty0response} and be atomized.
    \item \texttt{node\_\penalty0candidates} must be populated based on \texttt{verified\_\penalty0facts}; subjective guessing is not allowed.
    \item \texttt{candidate\_\penalty0sets} must not be arbitrarily spliced; explicit evidence chains from \texttt{verified\_\penalty0facts} must exist.
    \item If \texttt{graph\_\penalty0simplification} is not \texttt{null}, it must be extremely conservative; semantic changes are not allowed.
\end{enumerate}

\medskip
\textbf{[Output Format (strict JSON).]} The output JSON has four top-level parts.
\medskip
\noindent\textbf{(1) verified\_facts.}
\begin{tcblisting}{listing only, colback=appendixcodebg, colframe=appendixcodebg, boxrule=0pt, arc=0pt, left=0pt, right=0pt, top=0pt, bottom=0pt, breakable, listing options={style=appendixcode, basicstyle=\scriptsize\ttfamily, breaklines=true, breakatwhitespace=true, breakindent=0pt, aboveskip=0pt, belowskip=0pt, showstringspaces=false, columns=fullflexible, keepspaces=true}}
"verified_facts": [
  {"fact": "atomic-level fact",
   "evidence_quote": "verbatim snippet from a single tool_response"}
]
\end{tcblisting}
\medskip
\noindent\textbf{(2) node\_candidates.}
\begin{tcblisting}{listing only, colback=appendixcodebg, colframe=appendixcodebg, boxrule=0pt, arc=0pt, left=0pt, right=0pt, top=0pt, bottom=0pt, breakable, listing options={style=appendixcode, basicstyle=\scriptsize\ttfamily, breaklines=true, breakatwhitespace=true, breakindent=0pt, aboveskip=0pt, belowskip=0pt, showstringspaces=false, columns=fullflexible, keepspaces=true}}
"node_candidates": [
  {"node_id": "from query_graph.nodes[].id",
   "candidates": [
     {"name": "candidate entity name",
      "supporting_constraints": ["must be verbatim from query_graph.nodes[].constraints or query_graph.edges[].value"],
      "violating_constraints": ["must be verbatim from query_graph.nodes[].constraints or query_graph.edges[].value"],
\end{tcblisting}
\begin{tcblisting}{listing only, colback=appendixcodebg, colframe=appendixcodebg, boxrule=0pt, arc=0pt, left=0pt, right=0pt, top=0pt, bottom=0pt, breakable, listing options={style=appendixcode, basicstyle=\scriptsize\ttfamily, breaklines=true, breakatwhitespace=true, breakindent=0pt, aboveskip=0pt, belowskip=0pt, showstringspaces=false, columns=fullflexible, keepspaces=true}}
      "evidence_facts": ["must be exactly equal to a fact in verified_facts"],
      "confidence": "high | medium | low"}
   ]}
]
\end{tcblisting}
\medskip
\noindent\textbf{(3) solution\_check.}
\begin{tcblisting}{listing only, colback=appendixcodebg, colframe=appendixcodebg, boxrule=0pt, arc=0pt, left=0pt, right=0pt, top=0pt, bottom=0pt, breakable, listing options={style=appendixcode, basicstyle=\scriptsize\ttfamily, breaklines=true, breakatwhitespace=true, breakindent=0pt, aboveskip=0pt, belowskip=0pt, showstringspaces=false, columns=fullflexible, keepspaces=true}}
"solution_check": {
  "exists_fully_satisfying_candidate_set": true,
  "global_missing_constraints": ["..."],
  "candidate_sets": [
    {"mapping": {"node_id_1": "name_1", "node_id_2": "name_2"},
     "supported_constraints": ["..."],
     "missing_constraints": ["..."],
     "violating_constraints": ["..."],
     "confidence": "high | medium | low"}
  ],
  "best_candidate_set": {
    "mapping": {"node_id_1": "name_1"},
    "missing_constraints": ["..."],
    "confidence": "high | medium | low"},
  "reason": "..."
}
\end{tcblisting}
\medskip
\noindent\textbf{(4) graph\_simplification} (optional, rare cases; may be \texttt{null}).
\begin{tcblisting}{listing only, colback=appendixcodebg, colframe=appendixcodebg, boxrule=0pt, arc=0pt, left=0pt, right=0pt, top=0pt, bottom=0pt, breakable, listing options={style=appendixcode, basicstyle=\scriptsize\ttfamily, breaklines=true, breakatwhitespace=true, breakindent=0pt, aboveskip=0pt, belowskip=0pt, showstringspaces=false, columns=fullflexible, keepspaces=true}}
"graph_simplification": {
  "pruned_nodes": [{
    "node_id": "from query_graph.nodes[].id",
    "reason": "explain redundancy based on evidence"}],
  "merged_nodes": [{
    "target_node": "retained node_id",
    "merged_from": ["merged node_id1", "merged node_id2"],
    "reason": "explain semantic equivalence based on evidence"}],
    "pruned_edges": [{
    "edge_id": "from query_graph.edges[].id",
    "reason": "explain redundancy based on evidence"}],
  "merged_edges": [{
\end{tcblisting}
\begin{tcblisting}{listing only, colback=appendixcodebg, colframe=appendixcodebg, boxrule=0pt, arc=0pt, left=0pt, right=0pt, top=0pt, bottom=0pt, breakable, listing options={style=appendixcode, basicstyle=\scriptsize\ttfamily, breaklines=true, breakatwhitespace=true, breakindent=0pt, aboveskip=0pt, belowskip=0pt, showstringspaces=false, columns=fullflexible, keepspaces=true}}
    "target_edge": "from query_graph.edges id",
   "merged_from": ["merged edge_id1", "merged edge_id2"],
    "reason": "explain equivalence based on evidence"}]
}
\end{tcblisting}
\end{tcolorbox}

\subsection{Visit Tool Extraction Prompt}
The \texttt{visit} tool condenses a fetched webpage against the agent's current goal using the following extraction prompt, whose output feeds the evidence-driven state update $\Psi$ (\S3.3).
\begin{tcolorbox}[colback=appendixboxbg, colframe=appendixboxframe, boxrule=0.4pt, arc=2pt, left=6pt, right=6pt, top=4pt, bottom=4pt, title=Visit Tool Extraction Prompt, title after break={}, fonttitle=\bfseries\small, coltitle=white, colbacktitle=appendixboxtitle, breakable]
\small\sloppy
Please process the following webpage content and user goal to extract relevant information.

\medskip
\textbf{Webpage Content.} \texttt{\{webpage\_content\}}

\medskip
\textbf{User Goal.} \texttt{\{goal\}}

\medskip
\textbf{Task Guidelines.}
\begin{enumerate}\setlength{\itemsep}{2pt}\setlength{\parskip}{0pt}\setlength{\topsep}{0pt}
    \item \emph{Content scanning} (\texttt{rational}): locate the sections of the webpage that directly address the user's goal.
    \item \emph{Key extraction} (\texttt{evidence}): pull out the most relevant information, preserving the original context verbatim as far as possible (this may span more than three paragraphs); no important detail should be omitted.
    \item \emph{Summary output} (\texttt{summary}): distill the extracted content into a concise, logically ordered paragraph and assess its contribution toward the goal.
\end{enumerate}

\medskip
The final output is a JSON object with \texttt{rational}, \texttt{evidence}, and \texttt{summary} fields.
\end{tcolorbox}

\subsection{Forced Answer Prompt}
When the maximum context length is reached, the agent is forced to stop exploring and emit a best-effort answer via the following prompt.
\begin{tcolorbox}[colback=appendixboxbg, colframe=appendixboxframe, boxrule=0.4pt, arc=2pt, left=6pt, right=6pt, top=4pt, bottom=4pt, title=Forced Answer Prompt, title after break={}, fonttitle=\bfseries\small, coltitle=white, colbacktitle=appendixboxtitle, breakable]
\small\sloppy
You have now reached the maximum context length you can handle. You should stop making tool calls and, based on all the information above, think again and provide what you consider the most likely answer in the following format:\\
\texttt{<think>}your final thinking\texttt{</think>}

\medskip
\texttt{<answer>}your answer\texttt{</answer>}
\end{tcolorbox}

%% file: sections/AppendixH_CaseStudy.tex
\section{Case Study}
\label{app:case_study}

We present a complete G-ReAct trajectory that illustrates query-graph
initialization, graph-conditioned exploration, evidence-driven state updates,
and answer generation. The example is a BrowseComp-style deep-search question.
Each observation below is tied to an evidence record, so the state transition
and the final termination decision can be inspected directly.

\subsection{Question and Query Graph}

\noindent\textbf{Question.}\quad\textit{Identify the literary work referenced
by the following constraints. The work is a product of an author prolific in
the genre of suspense and terror. This author dedicated the work to a legally
recognized female life partner of the author and to the siblings of that
partner. Its conceptual origin was as a narrative complement to a distinct,
prior long-form prose fiction by the same creator, a link alluded to through a
shared depiction of a predictable, periodic darkening of a celestial body. This
prior work features a sequence where a character perceives, through
non-sensory means, the historical suffering of a younger female relative caused
by a male biological parent. The dedicatory phrasing intentionally recalls a
theoretical framework presented in a foundational socio-economic treatise from
the final decade of the 1800s, which analyzes the custom of a dominant gender
group claiming possession of individuals of another gender group as indicators
of social prestige. What two-word title designates the initial work?}

\paragraph{Query Graph $G_0$.}
The initializer $\phi$ parses the question into six abstract nodes and their
relations:

\begingroup
\renewcommand{\_}{\textunderscore\discretionary{}{}{}}
\raggedright
\begin{itemize}\setlength{\itemsep}{1pt}
    \item \texttt{Answer}: the two-word title to identify.
    \item \texttt{Author}: the work's author.
    \item \texttt{Prior\_Work}: a companion novel linked by the eclipse.
    \item \texttt{Socio-economic\_Treatise}: the foundational 1890s treatise.
    \item \texttt{Female\_Life\_Partner}: the author's legally recognized partner.
    \item \texttt{Character}: the character with the non-sensory perception.
\end{itemize}

\noindent The edges encode \texttt{has\_author} (Answer $\to$ Author),
\texttt{associated\_with} (the narrative complement, dedication, and shared
eclipse relations), and \texttt{related\_to} (the dedicatory clue and
\texttt{Socio-economic\_Treatise}).
\endgroup

\subsection{Graph-Conditioned Exploration}

At round $i$, the policy conditions on the fixed graph $G_0$, the preceding
state $S_{i-1}$, and the local interaction history $H_t^{(\mathrm{local})}$.
The rendered state supplies candidate sets, uncovered constraints, and the
accumulated fact pool $F_{\leq i-1}$ for the Think--Act--Observe cycle.

\begin{tcolorbox}[colback=appendixboxbg, colframe=appendixboxframe,
boxrule=0.4pt, arc=2pt, left=8pt, right=8pt, top=4pt, bottom=4pt,
before skip=3pt, after skip=3pt,
fonttitle=\bfseries\small, title=Round 1: Candidate Hypothesis and Work-Level Grounding,
coltitle=white, colbacktitle=appendixboxtitle, breakable]
\small
\textbf{Think.} Establish the candidate title and the shared-eclipse
connection before resolving the more delicate dedication allusion. \emph{Gerald's
Game} and \emph{Dolores Claiborne} are the leading candidate pair, but the
evidence policy must distinguish directly observed facts from an inference
about the wording of the dedication.

\smallskip
\textbf{Act.} The agent grounds the candidate pair through the official work
record:
\begin{tcblisting}{listing only, colback=appendixboxbg, colframe=appendixboxframe,
boxrule=0.3pt, arc=1pt, left=4pt, right=4pt, top=2pt, bottom=2pt,
listing options={style=appendixplaincode, basicstyle=\scriptsize\ttfamily, breaklines=true,
breakatwhitespace=true, breakindent=0pt, showstringspaces=false, columns=fullflexible, keepspaces=true, escapeinside={(*}{*)}}}
{
  "name": "search",
  "arguments": {
    "query": [
      "site:stephenking.com/(*\allowbreak*)works/novel/(*\allowbreak*)Gerald's Game Dolores Claiborne eclipse"
    ]
  }
}
\end{tcblisting}

\smallskip
\textbf{Observe $\to S_1$.} The official work record identifies \emph{Gerald's
Game}, Stephen King, and a Dolores reference during the eclipse. This grounds
the Answer, Author, and Prior\_Work hypotheses; the dedication text and
treatise clue remain open.
\end{tcolorbox}

\begin{tcolorbox}[colback=appendixboxbg, colframe=appendixboxframe,
boxrule=0.4pt, arc=2pt, left=8pt, right=8pt, top=4pt, bottom=4pt,
before skip=3pt, after skip=3pt,
fonttitle=\bfseries\small, title=Round 2: Parallel Evidence Collection,
title after break={},
coltitle=white, colbacktitle=appendixboxtitle, breakable]
\small
\textbf{Think.} Search the two independent unresolved branches together: the
dedication wording, and the companion-work / eclipse context. The agent keeps
the provenance of a bibliographical transcription separate from that of an
official work record.

\smallskip
\textbf{Act.} The agent sends a paired search request:
\begin{tcblisting}{listing only, colback=appendixboxbg, colframe=appendixboxframe,
boxrule=0.3pt, arc=1pt, left=4pt, right=4pt, top=2pt, bottom=2pt,
listing options={style=appendixplaincode, basicstyle=\scriptsize\ttfamily, breaklines=true,
breakatwhitespace=true, breakindent=0pt, showstringspaces=false, columns=fullflexible, keepspaces=true, escapeinside={(*}{*)}}}
{
  "name": "search",
  "arguments": {
    "query": [
      "Gerald's Game six good women dedication",
      "site:stephenking.com/(*\allowbreak*)works/novel/(*\allowbreak*)dolores-claiborne eclipse Selena"
    ]
  }
}
\end{tcblisting}

\smallskip
\textbf{Observe $\to S_2$.} The returned front-matter transcription supplies
the dedication wording, while the official record for \emph{Dolores Claiborne}
confirms the eclipse setting and the relevant characters. The Answer, Author,
Prior\_Work, Female\_Life\_Partner, and Character candidates now have
support; the treatise identity and the wording--treatise bridge need one final
resolution.
\end{tcolorbox}

\begin{tcolorbox}[colback=appendixboxbg, colframe=appendixboxframe,
boxrule=0.4pt, arc=2pt, left=8pt, right=8pt, top=4pt, bottom=4pt,
before skip=3pt, after skip=3pt,
fonttitle=\bfseries\small, title=Round 3: Dedication--Treatise Resolution,
title after break={},
coltitle=white, colbacktitle=appendixboxtitle, breakable]
\small
\textbf{Think.} The remaining relation connects the recorded dedication clue
to the candidate socio-economic treatise. The agent must resolve the treatise
identity while preserving whether the relation is directly observed or inferred.

\smallskip
\textbf{Act.} The agent visits the primary text of the candidate treatise:
\begin{tcblisting}{listing only, colback=appendixboxbg, colframe=appendixboxframe,
boxrule=0.3pt, arc=1pt, left=4pt, right=4pt, top=2pt, bottom=2pt,
listing options={style=appendixplaincode, basicstyle=\scriptsize\ttfamily, breaklines=true,
breakatwhitespace=true, breakindent=0pt, showstringspaces=false, columns=fullflexible, keepspaces=true, escapeinside={(*}{*)}}}
{
  "name": "visit",
  "arguments": {
    "url": [
      "https://www.gutenberg.org/(*\allowbreak*)files/833/(*\allowbreak*)833-h/833-h.htm"
    ],
    "goal": "Extract the 1899 theory of gendered(*\allowbreak*) possession and social prestige relevant to the dedication clue."
  }
}
\end{tcblisting}

\smallskip
\textbf{Observe $\to S_3$.} The primary text identifies Veblen's 1899 treatise
and its account of ownership, status, and gendered possession. It resolves
\texttt{Socio-economic\_Treatise} as \emph{The Theory of the Leisure Class}.
The state records the dedication--treatise bridge as an \emph{inferred}
relation supported by the recorded wording and retrieved theory, not as a
direct attribution of authorial intent. All six answer-identifying nodes now
have a consistent candidate.
\end{tcolorbox}

\subsection{State Update and Final Answer}

\paragraph{Evidence-Driven State Update ($\Psi$).}
Let $\tau_3$ denote the complete local trace from the third round. Following
the main-text transition, the update is
\[
S_3 \leftarrow \Psi(S_2, \tau_3)
= (F_{\leq 3},\; \mathcal{C}_3,\; \Sigma_3).
\]

\begin{itemize}\setlength{\itemsep}{1pt}
    \item $F_{\leq 3}$: evidence records for the official work pages, the
    dedication wording, and the primary treatise text.
    \item $\mathcal{C}_3$: a single consistent candidate assignment:
    \begin{itemize}\setlength{\itemsep}{0pt}
        \item \texttt{Answer} $\to$ \emph{Gerald's Game}
        \item \texttt{Author} $\to$ Stephen King
        \item \texttt{Prior\_Work} $\to$ \emph{Dolores Claiborne}
        \item \texttt{Socio-economic\_Treatise} $\to$ \emph{The Theory of the
        Leisure Class}
        \item \texttt{Female\_Life\_Partner} $\to$ Tabitha King
        \item \texttt{Character} $\to$ Dolores
    \end{itemize}
    \item $\Sigma_3^{\mathrm{miss}} = \emptyset$ and
    $\Sigma_3^{\mathrm{cands}}$ contains one compatible candidate combination.
    Its $\Sigma_3^{\mathrm{best}}$ assignment is the candidate set above; the
    dedication--treatise edge retains its \emph{inferred} provenance flag, and
    $\Sigma_3^{\mathrm{sat}} = \text{true}$ under the answer-resolution
    policy.
\end{itemize}

\paragraph{Final Answer.}
Upon $\Sigma_3^{\mathrm{sat}} = \text{true}$, the agent emits the answer
from \texttt{best\_candidate\_set}:

\begin{tcolorbox}[colback=appendixboxbg, colframe=appendixboxframe,
boxrule=0.5pt, arc=2pt, left=8pt, right=8pt, top=5pt, bottom=5pt,
title=Final answer, fonttitle=\bfseries\small, coltitle=white,
colbacktitle=appendixboxtitle]
\small\texttt{<answer>}\textbf{\emph{Gerald's Game}}\texttt{</answer>}
\end{tcolorbox}

Here, $\Sigma_3^{\mathrm{sat}} = \text{true}$ means that every
answer-identifying constraint has an evidence-supported candidate assignment
and no remaining incompatible candidate. It does \emph{not} convert the
inferred dedication--treatise bridge into a direct statement of authorial
intent. All six nodes in $G_0$ are filled, yielding a consistent answer
assignment.